\documentclass[notitlepage,12pt]{jedm}
\usepackage[table]{xcolor}
\usepackage{float}

\usepackage{url}
\let\bibhang\relax
\usepackage[natbibapa]{apacite}
\usepackage{subcaption}
\usepackage{hyperref}
\usepackage{graphicx}
\usepackage{tcolorbox}
\hypersetup{
  colorlinks   = true, 
  urlcolor     = blue, 
  linkcolor    = blue, 
  citecolor   = blue 
}
\shortcites{lin2025abcde}
\shortcites{smith2000conversation}
\shortcites{choi2005scaffolding}
\shortcites{kreijns2003identifying}
\shortcites{klonek2020capturing}
\shortcites{chang2023dramatic}
\shortcites{tay2002discourse}
\shortcites{qamarllms}
\shortcites{kojima2022large}
\shortcites{roberts2024needle}
\shortcites{ranganath2018understanding}

\newcommand{\best}[1]{\textcolor{blue}{\textbf{#1}}}

\begin{document}

\title{Leveraging Large Language Models to Identify Conversation Threads for  Collaborative Learning Analysis}
\date{} 


\newcommand{\authorFixedWidth}[1]{\parbox[t]{.25\textwidth}{\raggedright#1 \raisebox{0pt}[0pt][6pt]{}}}

\author{ \authorFixedWidth{{\large Prerna Ravi*}\\MIT CSAIL\\Cambridge, MA, USA\\prernar@mit.edu} \and 
\authorFixedWidth{{\large Dong Won Lee*}\\MIT Media Lab\\Cambridge, MA, USA\\dongwonl@mit.edu}  \and 
\authorFixedWidth{{\large Beatriz Flamia}\\Instituto Politécnico de Bragança\\Bragança, Portugal\\beatrizflamia@ipb.pt}   \and \authorFixedWidth{{\large Jasmine David}\\MIT STEP\\Cambridge, MA, USA\\jas.dav7654@gmail.com}   \and \authorFixedWidth{{\large Brandon Hanks}\\MIT STEP\\Cambridge, MA, USA\\bhanks@education.mit.edu}   \and \authorFixedWidth{{\large Cynthia Breazeal}\\MIT Media Lab\\Cambridge, MA, USA\\cynthiab@media.mit.edu}   \and \authorFixedWidth{{\large Emma Anderson}\\MIT STEP\\Cambridge, MA, USA\\eanderso@education.mit.edu } \and \authorFixedWidth{{\large Grace Lin}\\MIT STEP\\Cambridge, MA, USA\\gcl@mit.edu}}

\maketitle

\begin{abstract}
Understanding how ideas develop and flow in small-group conversations is critical for analyzing collaborative learning. A key structural feature of these interactions is threading—the way discourse talk naturally organizes into interwoven topical strands that evolve over time. While threading has been widely studied in asynchronous text settings, detecting threads in synchronous spoken dialogue remains challenging due to overlapping turns and implicit cues. At the same time, large language models (LLMs) show promise for automating discourse analysis but often struggle with long-context tasks that depend on tracing these conversational links. In this paper, we investigate whether explicit thread linkages can improve LLM-based coding of relational moves in group talk. We contribute a systematic guidebook for identifying threads in synchronous multi-party transcripts and benchmark different LLM prompting strategies for automated threading. We then test how threading influences performance on downstream coding of conversational analysis frameworks, that capture core collaborative actions such as agreeing, building, and eliciting. 
Our results show that providing clear conversational thread information improves LLM coding performance and underscores the heavy reliance of downstream analysis on well-structured dialogue. We also discuss practical trade-offs in time and cost, emphasizing where human-AI hybrid approaches can yield the best value. Together, this work advances methods for combining LLMs and robust conversational thread structures to make sense of complex, real-time group interactions.

{\parindent0pt
\textbf{Keywords:} Threading, Conversational Analysis, Large Language Models (LLMs), Collaborative Learning
}
\end{abstract}

\section{Introduction}
Analyzing how small groups collaborate through talk is a cornerstone of research in learning sciences, organizational behavior, and computational discourse analysis \citep{reyes2003supporting,bales1950interaction,armstrong2001individual,hennessy2016developing,cakir2005thread}. Decades of work have shown that the ways participants interact—how they build on, challenge, or elicit ideas—profoundly shape team outcomes, learning, and satisfaction \citep{smith1959effects,kim2008meeting}. However, tracing these interaction patterns reliably remains a persistent challenge, especially in \textbf{synchronous spoken setting}s where conversational dynamics are fluid, implicit, and overlapping \citep{mcdaniel1996identifying}.

A core structural feature of collaborative dialogue is \textbf{threading}, the idea that conversations naturally organize into interleaved topical strands that develop, branch, and reconnect over time \citep{mcdaniel1996identifying,ji2021responded}. Identifying these threads is vital for making sense of how ideas flow, how meaning is negotiated, and how groups reach shared understanding \citep{reyes2003supporting}. While threading is well-supported in asynchronous contexts like online forums—where reply structures are explicit and easily reviewable \citep{clark1991grounding}— its detection in real-time, multi-party speech is far more complex \citep{egbert1997schisming}. Prior work has highlighted that without reliable threading, important links between contributions can be lost, limiting the value of downstream conversational analysis frameworks that depend on these connections \citep{chang2023dramatic,herring1999interactional}.

In parallel, the rise of large language models (LLMs) has opened up new opportunities for automating conversation analysis \citep{lee2025capturing,martinenghi2024llms}. Yet, recent studies show that LLMs often struggle with tasks that require tracking long-range discourse dependencies \citep{hsieh2024ruler,liu2023lost}, such as classifying dialogue acts in extended transcripts. This raises a critical question: can explicit threading—when supplied to or generated by LLMs—improve the consistency and accuracy of their downstream coding?

This paper tackles this question in the context of collaborative learning analysis, using an example framework from \cite{lin2025abcde}.
We investigate how well LLMs can identify threads in synchronous small-group transcripts and whether providing explicit thread linkages—either human-annotated or LLM-generated—boosts the quality of relational coding for elicitation moves. In doing so, we bridge methodological gaps in discourse research with practical questions of cost, reliability, and human-AI complementarity.

Our contributions are threefold:
(1) We develop and test an operational definition and guidebook for threading in synchronous small-group talk, grounded in an iterative human annotation process,
(2) We benchmark multiple LLM prompting strategies for automated threading,
(3) We empirically test how threading impacts LLM performance on downstream coding of conversational analysis frameworks, providing new insights into where threads help, where errors persist, and what this means for scaling reliable collaborative discourse analysis.

Together, this work advances practical strategies for researchers and educators who want to use LLMs to make sense of complex group interactions while showing that better threading is not just a technical pre-processing step, but a critical scaffold for deeper conversational understanding.

\section{Related works}

\subsection{Understanding Interaction Patterns in Synchronous Collaboration}

Understanding how groups collaborate remains critical for advancing theories of collective work and for designing systems that support such interactions. Research has underscored that patterns of communication are among the strongest predictors of a work team’s success and are as influential as individual intelligence, personality, skill, or even the substantive content of discussions combined \citep{pentland2012new}. In particular, these communication patterns can reliably distinguish high-performing teams \citep{pentland2012new}. Scholars have also emphasized the importance of analyzing these fine-grained interaction patterns in the context of collaborative learning \citep{stahl2006group}. In the education space, conversational analysis provides a powerful lens for examining the social, communicative, and problem-solving interactions that underlie collaborative learning and group performance \citep{cakir2005thread}. Highlighting structures such as turn-taking conventions and adjacency pairs—typical sequences like question-answer or offer-response—helps reveal how participants negotiate meaning, coordinate actions, and maintain coherent dialogue \citep{smith2000conversation}. These techniques form the backbone of natural spoken interaction \citep{smith2000conversation}.

However, these natural multi-party settings are marked by spontaneous and overlapping talk, disfluencies, and unclear sentence boundaries—features that have presented scholars with significant challenges for analyzing speech over decades \citep{krauss1977role,shriberg2005}. To address this, researchers typically start by examining who speaks to whom moment by moment \citep{gatica2006analyzing}. Over extended interactions, this helps uncover additional patterns of recurring group dynamics such as floor control, dominance, and collective interest \citep{gatica2006analyzing}.

Taken together, these insights highlight that \textbf{making sense of collaborative group talk depends on tracing how ideas and actions connect across turns}. We explore this further in the next section through the concept of threading.

\subsection{Threading in Synchronous Group Interactions}

Threading, a core concept in the study of group interactions, refers to how conversations naturally organize into coherent topical strands that develop, diverge, and reconnect over time \citep{mcdaniel1996identifying}. Identifying these threads is vital for understanding how ideas flow and evolve, how meaning is negotiated, and how participants coordinate actions in collaborative settings \citep{reyes2003supporting}. Threads also serve as a prerequisite for more advanced methods such as social network analysis \citep{reyes2003supporting}. 
One practical application of threading is the detection of dominant participants, as they tend to both contribute more and receive more responses in conversation \citep{levine1990progress}. Identifying such patterns is valuable, since dominance is a key factor shaping a group’s social structure and overall dynamics \citep{kim2008meeting,desanctis1987foundation}.
Providing feedback on these dynamics can help groups adapt, leading to better collaboration and greater satisfaction \citep{smith1959effects}.

\cite{cakir2005thread} points out that much of the existing work on threading, however, has focused on asynchronous contexts—such as forums \citep{tay2002discourse}, email chains \citep{venolia2003understanding}, or discussion boards \citep{jeong2003sequential}—where clear textual markers, reply structures, or interface features make topic boundaries more explicit. In these settings, conversation data is easily reviewable \citep{clark1991grounding}, allowing researchers and systems alike to trace long-distance connections or nested replies. Yet, as \cite{ji2021responded} point out, even in online contexts, many methods overlook long-range dependencies \citep{wang2010making}, assuming utterances only respond to their immediate neighbors \citep{zhao-etal-2018-unsupervised,jiao2018find}.

In contrast, \textbf{threading in synchronous interactions is much more challenging and understudied.} Unlike written or online discussions, where explicit textual cues can signal breaks or topic shifts, threads in synchronous conversations tend to be ephemeral, implicit and fluid, making their identification especially complex \citep{egbert1997schisming}. Responses may occur after long stretches of unrelated talk, and contributions can overlap or interrupt each other, making it challenging to trace not only who is responding to whom, but where one topical strand ends and another begins \citep{mcdaniel1996identifying}. When the structure of a transcript fails to reflect how people naturally build on each other’s ideas, important connections can be obscured, making it harder to sustain meaningful dialogue and shared understanding \citep{reyes2003supporting}. Other scholars have similarly observed this misalignment between temporal flow and thread order \citep{herring1999interactional,davis2002cooperation}. \cite{ji2021responded} also notes that this ambiguity is especially pronounced in argumentative multi-party exchanges, where initiation–response pairs are more difficult to identify than in everyday talk. In practice, even trained researchers often disagree on how to segment discourse into threads, with reported agreements ranging from as low as 15\% to as high as 87\% when threading the same conversation \citep{mcdaniel1996identifying}. These disagreements highlight that threading is still as much an interpretive practice as it is a procedural one—requiring significant human time, judgment, and negotiation to resolve ambiguities in real conversational data.

Given the messiness of threading in synchronous settings, clear operational definitions are essential for threading to be reliable. Since early literature offered \textbf{very few concrete guidelines for how to identify threads in practice} \citep{grimes1975thread,black1983real,rose1995discourse}, \cite{mcdaniel1996identifying} sought to address this gap with a working definition, describing a thread as \textit{“a stream of conversation in which successive contributions continue a topic, following an initial contribution which introduces a new topic”}, and emphasizing the role of linguistic cues as signals for continuation. 
Since then, other researchers have built on this work. \cite{ji2021responded} shows that reliable thread detection also benefits from modeling deeper latent factors like topic alignment and discourse role dependencies—pushing beyond surface text features.

Taken together, these challenges underscore why our work aims not just to operationally define a thread but to develop a clear, systematic guidebook with examples for identifying threads in synchronous conversations. Ensuring consistency in human-coded threading is crucial: without strong human agreement, automated methods designed to reduce labor will likewise struggle for consistency. To motivate this, the next section reviews how recent research has attempted to automate thread detection—and why better human-grounded frameworks remain necessary.

\subsection{LLM-Based Conversation Analysis}

Recent advances in large language models (LLMs) have enabled promising approaches to automate the analysis of multi-party conversations. Across studies, researchers have explored how prompt design, segmentation strategy, and model capabilities affect performance on discourse-level tasks. While these efforts illuminate important challenges in conversational understanding, they stop short of addressing the role of threading as a structural unit—an area our work directly targets.

One line of work investigates how segmentation and prompting choices affect LLM performance. \cite{tran2024multi} found that using short, fixed-length windows (e.g., 5-turn) and simple binary classification prompts led to more consistent and accurate LLM outputs for scoring instructional quality in classroom discussions. Their results underscore how segmentation strategy shapes model performance—mirroring our interest in comparing sliding-window and thread-based segmentation for downstream coding tasks.

A complementary direction focuses on frameworks for automated annotation. \cite{petukhova2025fully} propose a fully automated pipeline for discourse annotation, pairing frequency-guided decision trees with GPT-4o to label utterances across taxonomies like Speech Functions and SWBD-DAMSL. Their method outperformed prior handcrafted schemes and human annotations, demonstrating the feasibility of structured, high-throughput annotation with LLMs. However, their pipeline remains grounded in turn-level analysis and does not model interleaved topical or interactional threads.

Efforts to capture latent discourse structures—such as collaborative reasoning—through prompting techniques also reveal the limits of current LLMs. For instance, \cite{lee2025capturing} introduce CoComTag, which uses prompts incorporating persona assignment, addressee prediction, and chain-of-thought reasoning to label collaborative competencies in student dialogues. While the approach achieved moderate agreement with human coders (kappa = 0.67), their analysis found that LLMs often relied on shallow cues (e.g., question form) and struggled with context disambiguation—a challenge we address through explicit thread modeling.

Further highlighting these limitations, \cite{qamarllms} identify core deficits in LLMs’ ability to model interactional dynamics, such as turn-taking and dialogue structure. Their study shows that performance on dialogue act classification is tightly coupled with these underlying competencies, and that LLMs frequently underperform compared to rule-based baselines. This suggests that enhancing structural awareness, such as tracking conversational threads, could provide the missing scaffolding for accurate dialogue interpretation.

\cite{martinenghi2024llms} reach a similar conclusion in their analysis of ChatGPT’s pragmatic capabilities in the STAC dataset. While ChatGPT performed well on some acts (e.g., Accept, Refuse), it struggled with others requiring deeper pragmatic inference and speaker modeling, such as Counteroffer. Their results again point to a reliance on surface heuristics over structured reasoning—reinforcing the need for richer contextual representations.

Taken together, these studies emphasize the importance of segmentation, structure, and context modeling in LLM-based conversation analysis. Yet, none directly engage with the concept of threading—the dynamic, interleaved structure through which small-group conversations unfold. Our work aims to address this gap.

\section{Contributions}
Building on prior literature, our work aims to advance how threading is conceptualized and used in the analysis of synchronous, small-group conversations. Despite positions that conversational threads are critical for understanding topical continuity and collaborative sense-making, there remains a lack of clear, operational definitions that can be reliably applied in real-time or retrospective analysis, especially in synchronous settings where implicit cues make threads difficult to trace.

The first goal of this paper is therefore to propose and test an operational definition of threads for synchronous, small-group conversations. By clarifying how a thread can be identified and represented in transcript data, we aim to offer a foundation for both human annotation and automated analysis of group conversations.

Our second goal is to examine whether large language models (LLMs) can identify thread structures from group conversation data. Specifically, we ask whether LLMs can identify and label threads  and how their performance compares to more conventional segmentation approaches, such as fixed sliding windows (where the transcript is divided into chunks of a set number of utterances to provide local context for each prediction).

Finally, we seek to understand whether explicitly providing thread labels to an LLM improves its effectiveness when performing deductive coding tasks on conversational transcripts. In other words, does pre-segmenting by thread help the model apply codes more consistently and accurately? And if so, what is the best way to supply these thread labels when prompting an LLM?

Together, these aims lead us to the following research questions:

\begin{itemize}
\setlength{\itemsep}{-0.2em}
    \item \textbf{RQ1:} How can threads be operationally defined in small-group synchronous conversations?
    \item \textbf{RQ2a:} Can LLMs identify and apply our thread definition reliably, and if so, what approaches could we use for such purposes?
    \begin{itemize}
        \item \textbf{RQ2b:} What specific elements of threading are LLMs adept at handling, and which present challenges?
    \end{itemize}
    \item \textbf{RQ3:} How does the performance of LLMs on deductive coding tasks (for Conversational Analysis) change when conversation transcripts include explicit thread labels compared to when they do not? 

\end{itemize}

We also examine the trade-offs between performance, time, and cost when using LLMs for conversational analysis tasks.
Through these questions, we aim to advance practical approaches for analyzing collaborative, multi-party conversations leveraging LLMs. Our findings can ultimately inform the development of tools that use LLMs to support qualitative researchers in their coding workflows, help teachers provide more meaningful formative feedback, and create new opportunities for students to engage in self-reflection.


\section{Preliminaries}
Understanding how conversational ideas persist, evolve, and interweave across turns is central to our study. In this section, we first define what we mean by a “thread” in synchronous small-group talk. Then,  next explore how threads can be used as an intermediate representation to support richer conversational analysis. By making the underlying connections between contributions explicit, threads offer a structured scaffold that can reduce ambiguity and effort in applying collaborative discourse coding schemes, and potentially improve both human and LLM performance on fine-grained interaction analysis.

\subsection{Threading}

\begin{figure}
    \centering
    \includegraphics[width=1\linewidth]{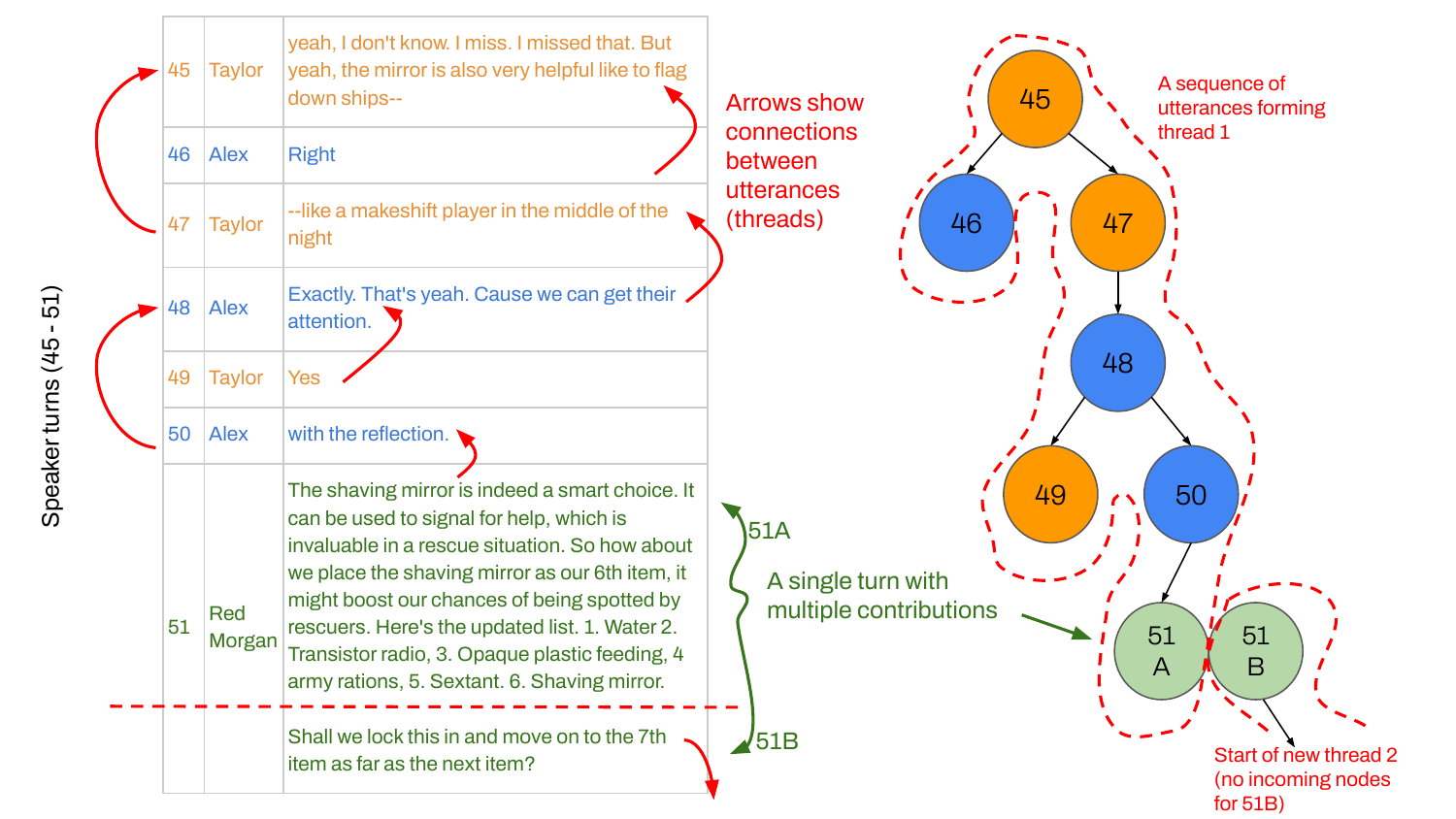}
    \caption{Anatomy of a Thread}
    \label{fig:thread_anatomy}
\end{figure}

Defining what constitutes a conversational thread is the essential first step in our approach. We use \cite{mcdaniel1996identifying}’s work as a starting point. A conversation thread is defined as a sequence of connected utterances that revolve around the same topic or theme. Each thread is sequentially organized across multiple speakers’ turns, with each turn contributing to the evolving topic. A \textit{turn} is an uninterrupted utterance by a single speaker, while a \textit{contribution} is the portion of a turn that belongs to a single thread—meaning that a single turn can contain multiple contributions if it references more than one idea or topic (See Figure \ref{fig:thread_anatomy}). 

While this definition provides a clear conceptual starting point, our experience shows that it is insufficient on its own for consistent annotation in real-world, synchronous small-group talk. Such conversations often feature overlapping talk, subtle topic shifts, and interruptions, all of which create ambiguity in determining thread boundaries.

To address these challenges, we developed a threading framework with explicit operational rules, decision criteria, and illustrative examples. This guidebook equips annotators to handle edge cases and conversational nuance consistently 
to support reliable, replicable annotation. We present this framework in Section \ref{threading-framework} and describe how it was subsequently used for human and LLM annotation.

\subsection{Downstream Conversational Analysis}
\label{CA_main}

After establishing methods to detect conversational threads with reasonable reliability, we turn to asking how useful these threads are for supporting downstream collaborative discourse analysis. 
Many coding schemes for collaborative discourse have shown dependency on identifying how one contribution relates to another \citep{schermuly2012discussion,hennessy2016developing,bales1950interaction}. Yet in synchronous conversations, these connections often remain implicit or buried within overlapping turns and fluid topic shifts, forcing coders to expend significant time and judgment to reconstruct them. While humans can do this with training, relying solely on manual reconstruction is resource-intensive and hard to scale. 

Given this, it remains an open question whether LLMs—like human coders—also need access to these explicit thread linkages to perform fine-grained collaborative coding reliably. In section \ref{ABCDE-intro}, we test this hypothesis by conducting a case study with the ABCDE framework \citep{lin2025abcde}, an example of a conversational analysis well suited to this question. This framework highlights relational moves like agreement, elaboration, and elicitation—each of which depends on correctly identifying conversational links across turns. By examining how LLM performance changes when provided with explicit threads, we build on our prior results to assess how thread awareness shapes the accuracy and consistency of automated discourse analysis.

\section{Dataset}
\begin{figure}
    \centering
    \includegraphics[width=0.75\linewidth]{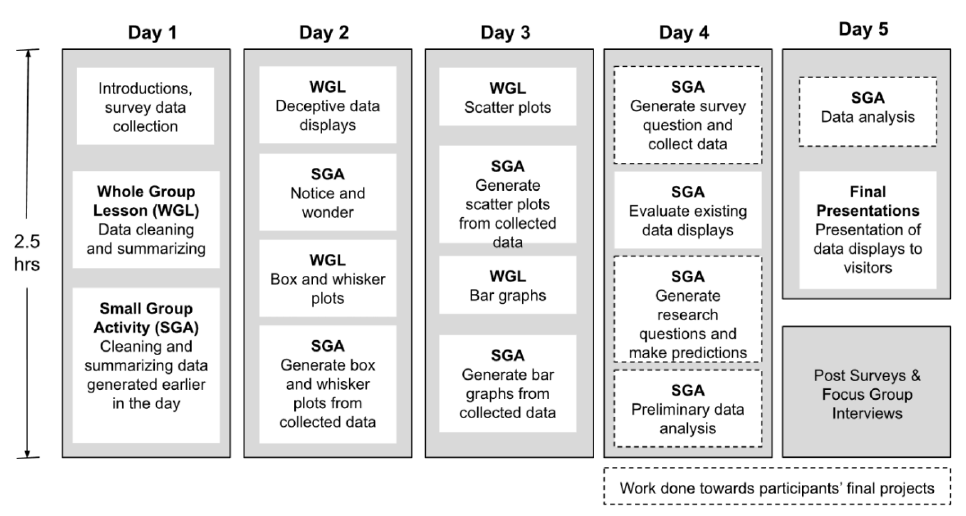}
    \caption{Activities performed by students in the data science workshop spanning our dataset}
    \label{fig:activities}
\end{figure}

In this section, we describe the dataset and annotation process underpinning this paper's threading and downstream conversational analysis tasks. 

Our dataset draws on two distinct small-group collaboration scenarios. The first comes from a five-day virtual data science workshop held in February 2024 with high school students. We analyzed transcripts from six of the nine students who attended the workshop. The students, ranging from 10-12th grade, represented diverse racial identities and had varying levels of math experience—from honors students to those reporting math anxiety and limited confidence with data science. Throughout the week, students collaborated in small groups to analyze data, create visualizations in Google Sheets, generate original research questions, and share their findings (see Figure \ref{fig:activities}). For these activities, the six students were divided into two groups of three, each working alongside an AI peer agent named Oscar.  The second scenario involves a team consensus-building exercise in a face-to-face setting with two adults and an AI-powered conversational agent named Red Morgan. In this task, participants collaboratively ranked items to salvage during a sinking ship survival scenario, following the classic team-building exercise described by \cite{johnson1991joining}. Both studies were approved by our institution's IRB. The group discussions were recorded, transcribed, and cleaned for further qualitative coding and analysis. All names used in this paper are pseudonyms.

\subsection{\textbf{Threading}: Guidebook Development and Validation}
\label{threading-framework}

We developed a framework for reliably detecting threads in synchronous, small group conversations. We describe its development process below. 

\paragraph{Step-by-Step Guide for Identifying Threads}

The guidebook described below was created as an outcome of an iterative social moderation process \citep{shaffer2017quantitative} conducted by three trained researchers, through which annotation criteria were refined and edge cases were systematically resolved. We describe this process in greater detail at the end of this section. The guidebook instructs coders to:

\begin{itemize}
\setlength{\itemsep}{-0.2em}
    \item Treat the first contribution as the start of the first thread.
    \item Classify each subsequent contribution by determining whether it replies to or extends a prior contribution. If it does, it continues the same thread.
    \item If the contribution neither responds to nor builds on a prior utterance, it begins a new thread.
    \item A \textit{reply} is defined as any contribution that answers a question, clarifies an issue, or adds directly to the prior idea.
\end{itemize}
Inspired by \cite{chang2023dramatic}, as a foundational rule, coders are instructed to start with the immediate previous utterance and work backward, eliminating candidates one by one until they find the closest appropriate link for the current line. This backwards-checking process should be limited to a window of \textit{N} prior utterances; if no link is found within that window, the contribution is marked as the start of a new thread. This window size \textit{N} depends on the context of the conversation and the nature of the task speakers are communicating.

When coding transcripts that include conversational AI agents (as was the case with our dataset introduced earlier in this section), the guidebook reminds coders to account for possible response delays or interruptions from the agent due to technological limitations. For example, an agent’s utterance might respond to an idea that was raised several turns earlier, requiring extra care to correctly link its contribution to the appropriate thread.

Finally, the guidebook also notes that what constitutes a \textit{new topic}—and therefore a new thread—must be defined in relation to the specific context of the dataset’s conversation, often requiring semantic interpretation and discussion among coders to reach a shared understanding.

\paragraph{Guidebook: Specific Scenarios}

The guidebook outlines specific scenarios that require careful judgment. Given the inherently messy nature of synchronous conversations, specifying these scenarios helped clarify ambiguous thread boundaries and achieve greater consistency in annotation decisions. 

\begin{figure}
  \centering
  \begin{subfigure}{0.95\linewidth}
    \centering
    \includegraphics[width=\linewidth]{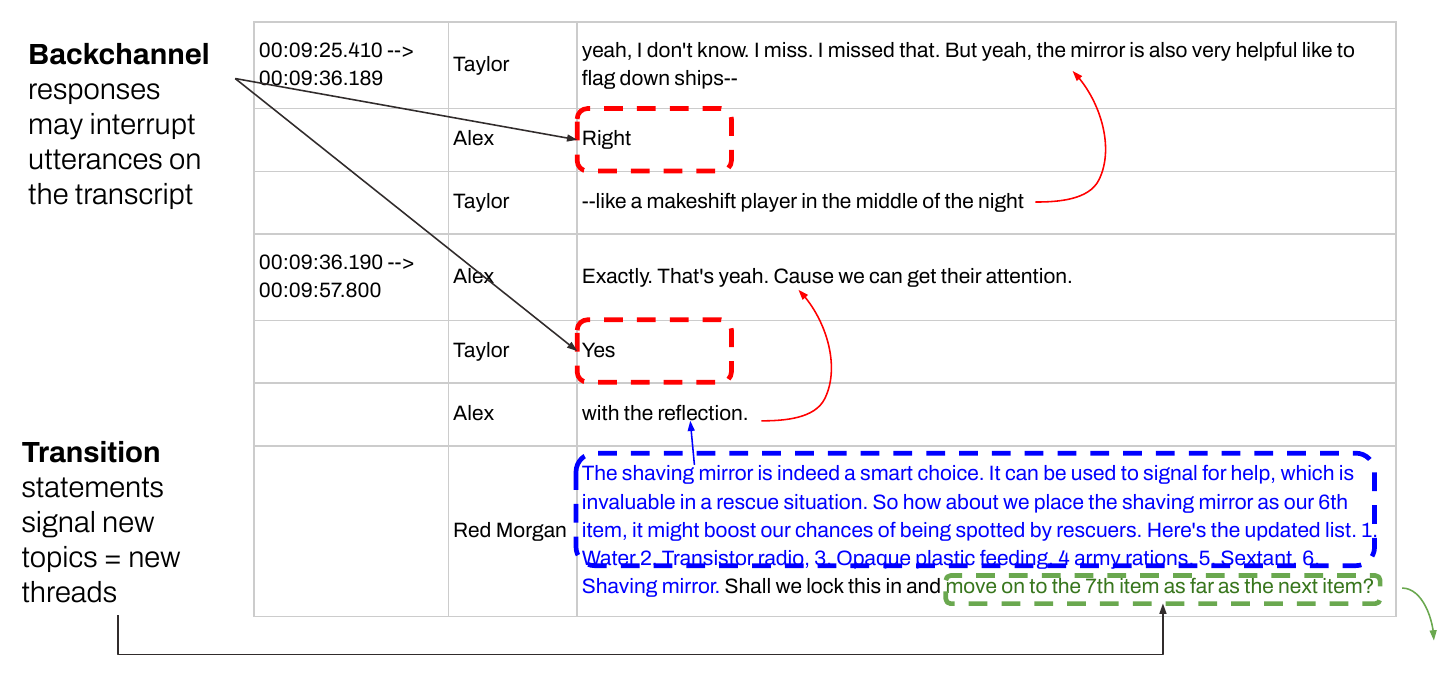}
    \caption{Examples of threads with backchannel responses and transition statements.}
    \label{fig:threading_guidebook_2}
  \end{subfigure}
  
  \vspace{0.5cm} 

    \begin{subfigure}{0.95\linewidth}
    \centering
    \includegraphics[width=\linewidth]{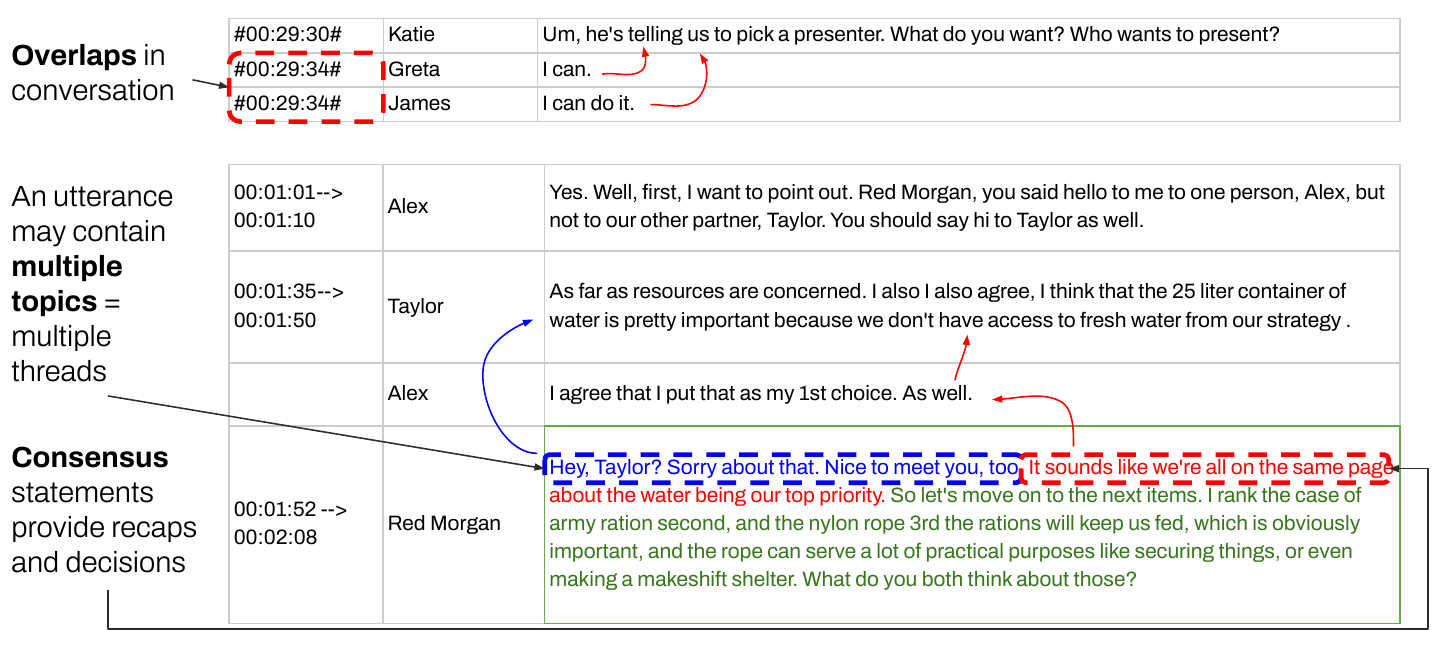}
    \caption{Examples of threads with overlaps, multiple contributions, and consensus statements.}
    \label{fig:threading_guidebook_1}
  \end{subfigure}
  
  \caption{Illustrative examples of conversation threads showing overlaps, multiple contributions within a single turn, consensus, backchannels, and transitions.}
  \label{fig:threading_guidebook_combined}
\end{figure}

\textbf{Backchannel responses.} Very brief acknowledgments (e.g.,\textit{ “yeah”, “hmm”}) do not constitute substantive contributions and should not break the continuity of the original speaker’s thought. For example, if a speaker resumes their statement after an interruption by a backchannel, their new utterance should link back to their original line, not the acknowledgment. See Figure \ref{fig:threading_guidebook_2} for a concrete example. 

\textbf{Transition statements.} Phrases that clearly signal a topic shift (e.g., \textit{“Let’s move on to the next problem”}) mark the beginning of a new thread. Coders are encouraged to look for keywords and discourse markers that indicate such transitions. See Figure \ref{fig:threading_guidebook_2} for a concrete example. 

\textbf{Split contributions.} Occasionally, a single utterance may contain multiple ideas that link to different parts of the conversation. In these cases, coders split the contribution and assign each portion to its appropriate thread. For example, a sentence that both summarizes previous discussion and proposes a new topic will be coded with two thread identifiers. See Figure \ref{fig:threading_guidebook_1} for a concrete example. 

\textbf{Consensus statements.} Summaries or clarifications that reflect a group’s collective decision should be linked back to the most recent relevant utterance. This ensures that the evolving group understanding is properly traced. See Figure \ref{fig:threading_guidebook_1} for a concrete example. 

\textbf{Use of timestamps.} When multiple speakers contribute simultaneously or nearly so, timestamps help coders discern which utterances are linked. This is especially important in high-tempo group discussions with overlapping turns. See Figure \ref{fig:threading_guidebook_1} for a concrete example. 

\textbf{Multiple plausible links.} If two equally reasonable candidate utterances could serve as the link, coders are instructed to choose the one closest in distance to the current line.

\textbf{Direct address vs. true linkage.} Addressing another speaker by name does not automatically mean that the contribution responds to that person’s previous utterance. Coders are reminded to determine the true topical link based on the content, not just conversational cues.


\paragraph{Annotation Validation}
\label{threading_human_annotation}
The human coding team consisted of three trained qualitative researchers, none of whom had participated in the original data collection for the transcripts used in this study. Such separation reduces potential bias and ensures that coders relied solely on the conversation’s transcript content and the operational definitions provided. In order to refine the guidebook and reach consistent interpretations of threads, all three researchers independently annotated the same subset of utterances with thread linkages across multiple iteration rounds. These utterances were drawn from four different transcripts covering both collaborative scenarios described earlier. As the process was iterative, discrepancies and points of confusion were systematically discussed and resolved using a social moderation process \citep{shaffer2017quantitative}. Emerging ambiguities and new edge cases identified during this process were added to the guidebook to support consistent decision-making throughout the remaining coding process. Annotators refined the threading guidebook over the course of coding 575 utterances in total across four transcripts. Inter-coder reliability, measured using Cohen’s Kappa on the final (fourth) transcript of 200 utterances, reached \textbf{0.89}, indicating substantial agreement. After establishing this reliability threshold, two of the three researchers independently annotated the remaining transcripts, splitting the data evenly.  These carefully moderated annotations served as the primary ground-truth reference for evaluating the performance of LLMs on the conversational threading task.

\subsection{Downstream Conversational Analysis: Guidebook Development and Validation}
\label{ABCDE-intro}
Our study first evaluates how well LLMs can identify conversational threads, and then examines the value of these threads as a resource for conversational analysis. To assess their utility, we focus on an example of a conversational analysis task that we believe will be of interest to the broader research community.

 The \textbf{ABCDE framework} \citep{lin2025abcde} investigates the ways group members connect their ideas and contributions. To do so, it uses as an example of a practical, action-oriented coding scheme for collaborative talk. The ABCDE framework identifies five core relationally based categories of actions that are essential for productive small-group discussion:

\begin{enumerate}
\setlength{\itemsep}{-0.2em}
    \item \textbf{A} — \textit{Agreeing} with others’ ideas, highlighting explicit statements of concurrence that help build common ground and maintain group cohesion.
    \item \textbf{B} — \textit{Building on} others’ contributions by adding new information, elaborating on points, or extending reasoning—key practices for deepening shared understanding.
    \item \textbf{C} — \textit{Chatting} or engaging in informal social exchanges, which may appear off-task but are vital for trust and relationship-building in teams.
    \item \textbf{D} — \textit{Differing perspectives} by introducing disagreements or alternative viewpoints, which can stimulate critical thinking, prevent groupthink, and promote more informed decision-making.
    \item \textbf{E} — \textit{Eliciting responses or actions} from others through questions or prompts that invite wider participation, ensuring that diverse perspectives are included in the group’s sense-making process.
\end{enumerate}
As demonstrated by \cite{lin2025abcde}, this framework has been applied to varied collaborative settings—from high-school math activities to adult team-building tasks—to reveal distinct patterns. Notably, each code in the ABCDE framework depends on correctly identifying how one utterance relates to another. For example, determining whether a speaker’s contribution is extending their own earlier ideas or that of someone else’s requires understanding the interaction between turns. This dependency on inter-utterance connections means that frameworks like ABCDE \textbf{cannot be reasonably coded without threading}, i.e. making the links between utterances explicit. 

\paragraph{Annotation Validation} We employed a similar social moderation process \citep{shaffer2017quantitative} to section \ref{threading_human_annotation} to develop reliable ground-truth annotations for the ABCDE framework, drawing on the original codebook developed by \cite{lin2025abcde}. The same three trained researchers deductively applied the ABCDE codes across three transcripts (on a total of 478 utterances), working independently at first and then collaboratively resolving discrepancies. 
Despite the attempt to resolve the discrepancies in 478 utterances, the researchers were unable to reach acceptable reliability on all codes. Acceptable human inter-coder agreement was reached only on codes C, D, E with kappas of 0.75, 0.66, and 0.881 simultaneously (Table \ref{tab:human_ABCDE}). Since our goal was to provide a targeted example illustrating the effect of threading on downstream conversational analysis, we restricted subsequent analyses to the code with highest inter-coder reliability: \textbf{E \textit{(Elicit)}}. This choice ensured the use of robust human ground-truth labels. The same two researchers who completed the threading annotations then applied the validated E codes to the remaining split and threaded transcripts. 

\begin{table}[H]
    \centering
    
    \begin{tabular}{|c|c|c|c|c|c|}
        \hline
        Code & A: Agree & B: Build & \textbf{C: Chat} & \textbf{D: Differ} & \textcolor{blue}{\textbf{E: Elicit}} \\ \hline
        Kappa & 0.321 & 0.489 & \textbf{0.757} & \textbf{0.663} & \textcolor{blue}{\textbf{0.881}} \\ \hline
    \end{tabular}
    \caption{ABCDE Human–Human Interrator Reliability, \textbf{\textcolor{blue}{Blue}} indicates the categories which resulted in best interrator reliability: \emph{E: Eliciting responses or actions} \textbf{Bold} indicates categories with acceptable interrator reliability \emph{C: \textit{Chat}, and D: \textit{Differ} }}
    \label{tab:human_ABCDE}
\end{table}



In light of the aforementioned reliability issue and to streamline the scope of this paper, we focus our LLM experiments (described in subsequent sections) on \textbf{E (\textit{Elicit})}, which captures conversational moves aimed at drawing peers into the discussion, soliciting input, or prompting action. In addition to its near perfect coding agreement (highest kappa of 0.88), we selected E for its central role in collaborative learning dynamics: eliciting participation is directly tied to equitable turn-taking, idea diversity, and group cohesion \citep{lin2025abcde}.  Importantly, variants of the E code—such as asking clarifying or elaborative questions \citep{choi2005scaffolding}, inviting peers to share perspectives \citep{chin2010students}, or  consensus-checking \citep{stoeckel2024strategies}—appear in many other discourse analysis frameworks \citep{bales1950interaction,klonek2020capturing,hennessy2025analysing}, underscoring its recognition as a fundamental conversational behavior for fostering inclusion, deepening dialogue, and co-constructing knowledge across contexts.

\begin{table}[H]
\centering
\caption{Corpus statistics for the 12 conversation transcripts in our dataset.}
\renewcommand{\arraystretch}{1.2}
\begin{tabular}{|l|r|}
\hline
\textbf{Statistic} & \textbf{Value} \\\hline
Total utterances & 2,365 \\
Total words & 31,535 \\
Average utterances per dialogue & 197.08 \\
Average words per dialogue & 2,627.92 \\
Average words per utterance & 13.33 \\
No-thread utterances & 1,325 \\
Average turns between thread links & 1.28 \\
Min / Max turns between thread links & 0 / 13 \\\hline
\multicolumn{2}{|l|}{\textbf{Label proportions (ABCDE codes)}} \\\hline
A (Agree) & 6.93\% \\
B (Build on) & 11.25\% \\
C (Chat/Comment) & 22.45\% \\
D (Differing perspective) & 2.71\% \\
\textbf{E (Elicit response)} & \textbf{28.84}\% \\\hline
\end{tabular}
\label{tab:corpus_stats}
\end{table}

\subsection{Final Dataset}

Our raw corpus consists (Table \ref{tab:corpus_stats}) of 2,365 utterances drawn from 12 conversation transcripts spanning the two collaborative group contexts described above. Each transcript involves at least two human participants and an AI peer agent, with the majority comprising three human participants and the AI agent. Conversations were segmented at the utterance level, where each segment corresponds to a change in speaker. Each row in the dataset represents a single utterance and includes metadata such as the start timestamp, speaker ID, utterance transcript, and utterance index.

Across the corpus, dialogues average 197.1 utterances and 2,627.9 words per transcript, with utterances averaging 13.33 words each. In total, the dataset contains 31,535 words. Of the utterances, 1,325 are not linked to any thread (i.e., no-thread cases). Threaded utterances exhibit a mean gap of 1.28 turns between linked utterances, ranging from consecutive turns to a maximum separation of 13 turns.

For downstream conversational analysis, each utterance is annotated with one or more ABCDE codes, with the following proportions observed: A (Agree) – 6.93\%, B (Build on) – 11.25\%, C (Chat/Comment) – 22.45\%, D (Differing perspective) – 2.71\%, and \textbf{E (Elicit response) – 28.84\%}.

\section{Methods}
As with earlier sections, we split the methods section below for the LLM annotations into the threading and downstream CA tasks. 

\subsection{Prompting Large Language Models (LLMs) for Threading}
\label{threading_methods}

\begin{table}
\caption{Prompt structure for LLM threading experiments: The prompt for the LLM threading experiments was designed with several key components to ensure interpretability and alignment with the human-coded guidebook.}
\centering

\begin{tabular}{|>{\raggedright\arraybackslash}p{0.3\linewidth}|>{\raggedright\arraybackslash}p{0.7\linewidth}|}
\hline
\textbf{Component} & \textbf{Description} \\
\hline
\multicolumn{2}{|c|}{\textbf{Base Prompt (Zero-shot)}} \\
\hline
Persona& The LLM is instructed to adopt the role of an \textit{expert qualitative coder}, explicitly framing its task as labeling threads within a conversational transcript. \\
\hline
Dataset Context& The prompt clarifies that the data come from synchronous, small-group conversations, setting expectations for informal, overlapping, or fragmented speech. \\
\hline
Operational Definition& Provides a precise, shared definition of what constitutes a thread: a sequence of connected utterances on the same theme, organized across speaker turns. It clarifies that a single turn can contain multiple contributions, ensuring that complex or multi-topic utterances are handled accordingly. \\
\hline
Core Instructions& Step-by-step instructions outline how to traverse the transcript: starting with the immediate prior utterance, working backward, and identifying replies or new topics based on content. \\
\hline
Threading Codebook& The prompt includes eight edge-case rules adapted from the team’s threading guidebook. These cover special discourse patterns such as backchannel responses, topic transitions, consensus statements, self-continuations, and ambiguous references. Each rule is illustrated with concrete examples. \\
\hline
Output Format Specification& The prompt explicitly specifies the required output structure for consistency: a single line indicating the utterance number, speaker, and respond\_line (thread label). The model is told not to provide any extra explanation in its outputs. \\
\hline
\multicolumn{2}{|c|}{\textbf{Additions for Sliding Window}} \\
\hline
Incremental Labeling within Transcript Windows& In the \textbf{sliding window approach}, the prompt treats threading as an incremental task: the LLM is asked to generate a label for only the final utterance in each window of N utterances, rather than receiving and labeling the entire transcript at once as in the base prompt condition. The window includes the thread labels for the previous N–1 utterances, giving the LLM access to prior linking decisions as additional context for predicting the current thread label. \\
\hline

\end{tabular}
\label{tab:prompt-threading}

\end{table}

All large language model (LLM) experiments in this study were conducted using a range of \textbf{OpenAI models}, accessed via the OpenAI API. The models tested include \texttt{o3-mini}, \texttt{o1-mini}, \texttt{gpt-4.1}, \texttt{gpt-4o}, \texttt{gpt-4.1-mini}, \texttt{gpt-4.1-nano}, and \texttt{gpt-4o-mini}. We chose OpenAI models for their ease of use and the diversity of their reasoning capabilities. While newer reasoning‑focused models (e.g., OpenAI’s o3 and o3‑mini) have demonstrated superior performance on several reasoning tasks, such as scoring 87.3\% on the AIME (advanced mathematical problem-solving,) \citep{hendrycksmeasuring} with high effort and attaining 71.7\% on SWE‑bench (code understanding) \citep{jimenez2023swe} (versus 48.9\% for o1), these gains don’t necessarily generalize across novel domains. For instance, GPT‑4.1‑mini, while less accurate on raw coding benchmarks, excels in instruction following, latency, and long‑context handling—demonstrating $\approx45.1\%$ on hard instruction following, comparable to o3‑mini’s 50.0\%, and substantially reducing costs (83\%) while improving responsiveness. Additionally, some benchmarks still favor GPT‑4.1, on the MMLU benchmark \citep{hendrycksmeasuring}, GPT‑4.1 beats o3‑mini(high), achieving 90.2\% versus 86.9\%. These findings underline the need to empirically verify whether reasoning performance observed in established domains (like STEM or coding) holds true within our specific domain of conversation threading and education. We must ensure that gains in traditional benchmarks translate to real-world utility in our context before selecting a base model.

We investigate common prompting strategies in treading and identify how these strategies affect performance in the task of conversational threading in synchronous group transcripts, we test the following three distinct approaches commonly used in prior works: (1) All-At-Once Zero-Shot Prompting: a zero‑shot technique where the model receives a complete instruction and transcript for the task without any illustrative examples \citep{kojima2022large}, (2) All-At-Once Few-Shot Prompting: a few‑shot method where the entire prompt includes several input–output demonstrations (“shots”) along with the task instructions, enabling the model to learn from these examples to generalize to new cases \citep{brown2020language}, (3) Sliding Window Zero-Shot Prompting: a strategy typically used in scenarios requiring long-context handling (e.g., long documents, conversation threads) where the model is prompted on sequential “windows” or slices of context in a zero-shot manner—no examples are provided, and each window is treated independently without reference to prior windows \citep{sun2023chatgpt}. We provide more details below.


\textbf{All-at-once (1) Zero-Shot / (2) Few-Shot Prompting:} In this baseline condition, the LLM receives the entire transcript at once, along with the complete threading definition and codebook instructions. In the zero-shot variant, no additional examples are shown. For the one-shot and three-shot setups, the model is provided with one to three fully threaded transcripts alongside the prompt so it can infer patterns in how threads emerge and how conversational turns link together, before applying this understanding to a new, unlabeled transcript. 

\textbf{(3) Sliding Window Zero-Shot Prompting: }In the sliding window condition, the prompt frames threading as an \textit{incremental task}. The LLM is shown a fixed-size window of the transcript and asked to label only the final utterance within that window. By varying the window size, we tested how much immediate conversational context affects performance. Specifically, we experimented with three window sizes (\textbf{10}, \textbf{20}, and \textbf{30} utterances) to balance context coverage and prompt length constraints. We include additional \textit{thread labels} for the previous \textbf{N - 1} utterances within each window. By exposing the model to the threading decisions made so far, we test whether the LLM can reason more accurately about the next link by using prior labeling as explicit context. This mirrors how human coders often rely on surrounding labeled connections to make more consistent judgments about topic flow and thread continuity.



\subsection{Prompting LLMs for Downstream Conversational Analysis}

For the downstream conversational analysis studying the effects of the inclusion of threading, we focused on \texttt{gpt-4.1}. Our aim in the downstream conversational analysis task was not to identify the “best” model for the case study. In threading, model comparison was necessary because the task is novel and we wanted to understand how different models perform. In contrast, for the downstream task our primary interest was in evaluating the effect of providing threading as additional context to the LLM, holding the model fixed, rather than in comparing across models.

Here, we compare conditions where the model receives (a) the \textbf{raw transcript only} versus (b) the \textbf{transcript with thread labels}, where the thread labels come either from human annotations or from the best-performing LLM threading approach (Table~\ref{tab:threading-sliding_vary}). To the best of our knowledge, this is the first work to explicitly evaluate how providing explicit conversation structure, in the form of threading, affects a downstream collaborative coding task (such as labeling for E from the ABCDE framework). By integrating thread information directly into the prompt, we move beyond treating threading as an isolated prediction task and instead test its practical utility for improving higher-level conversational analysis.

\begin{table}[H]
\caption{Prompt structure for ABCDE Coding experiments}
\centering

\begin{tabular}{|>{\raggedright\arraybackslash}p{0.3\linewidth}|>{\raggedright\arraybackslash}p{0.7\linewidth}|}
\hline
\textbf{Component} & \textbf{Description} \\
\hline
\multicolumn{2}{|c|}{\textbf{Base Prompt (Zero-shot)}} \\
\hline
Persona& The LLM is instructed to adopt the role of an \textit{expert qualitative coder}, explicitly framing its task for deductive coding within a conversational transcript.\\
\hline
Dataset Context& The prompt clarifies that the data come from synchronous, small-group conversations, setting expectations for informal, overlapping, or fragmented speech. \\
\hline
ABCDE Codebook& The prompt includes clear defintions and examples of each of the five codes from the ABCDE framework.\\
\hline
Output Format Specification& The prompt explicitly specifies the required output structure for consistency: a single line indicating the utterance number, speaker, and an array with the codes (e.g. [], [A], or [B, D]). The model is told not to provide any extra explanation in its outputs.\\
\hline
\multicolumn{2}{|c|}{\textbf{Additions for Sliding Window with and without threads}} \\
\hline
Incremental Labeling within Transcript Windows& In the \textbf{sliding window approach}, the LLM is asked to generate a label for only the final utterance in each window of N utterances, rather than receiving and labeling the entire transcript at once as in the base prompt condition. In the \textbf{sliding window set-up with threads}, the window includes the thread labels for the N utterances, giving the LLM access to utterance links as additional context for predicting the current ABCDE label.\\
\hline

\end{tabular}

\end{table}

The prompt for ABCDE coding was designed to align with the human codebook and ensure interpretability.  
Full prompt templates are provided in Appendix~\ref{app:prompts}.  
Below, we outline the prompting strategies, grouped by whether the LLM receives the entire transcript (\textit{All-At-Once}) or a limited local context (\textit{Sliding Window}), and whether the transcript includes explicit thread labels.

\textbf{All-at-once Prompting (No Threads / Human Threads / LLM Threads):}  
In these baseline conditions, the LLM receives the \textit{entire transcript at once}, along with the complete ABCDE definitions and coding instructions.  
We vary whether the transcript includes no thread labels, human-provided thread labels, or LLM-generated thread labels from our best-performing threading approach (Table~\ref{tab:threading-sliding_vary}).  

\begin{itemize}
    \item \textbf{No Threads:} The model sees only the raw transcript and ABCDE definitions. This tests baseline coding performance without explicit conversational structure.
    \item \textbf{Human Threads:} The transcript is augmented with accurate human-annotated thread labels. This serves as an upper bound for how much high-quality structure can improve ABCDE coding.
    \item \textbf{LLM Threads:} The transcript includes thread labels produced by the best-performing LLM threading model. This tests whether automatically inferred structure can enhance coding compared to no threads.
\end{itemize}

\textbf{Sliding Window Prompting (No Threads / Human Threads / LLM Threads):}  
In these conditions, the prompt frames ABCDE coding as an \textit{incremental task}. The LLM is shown a fixed-size window of the transcript and asked to label only the \textit{final} utterance in that window.  
We use a window size of \textbf{10 utterances}, chosen based on threading experiment results, and vary whether the transcript includes no threads, human threads, or LLM threads.

\begin{itemize}
    \item \textbf{No Threads:} The model sees a raw transcript window without any thread labels. This tests whether local conversational context alone supports accurate coding.
    \item \textbf{Human Threads:} The transcript window includes human-provided thread labels for all $N-1$ preceding utterances. This tests whether accurate local structure improves coding in a constrained context.
    \item \textbf{LLM Threads:} The transcript window includes LLM-generated thread labels for the $N-1$ preceding utterances. This tests whether local, automatically inferred structure can substitute for human annotations.
\end{itemize}

\section{Experiments}
In this section, we describe the process we used evaluate results obtained from the LLM Prompting approaches above for threading and conversational analysis. 

\subsection{Threading Evaluation}
\label{threading_analysis}
First, we assess the overall performance of our different LLM threading approaches using three commonly used metrics by comparing the LLM-generated outputs from our different methods against the human-annotated ground truth labels: \textbf{accuracy} (quantifies the proportion of correct predictions across all instances), \textbf{macro-average F1-score} (provides a balanced measure of performance across codes), and \textbf{Cohen’s kappa} (inter-rater reliability, correcting for chance agreement– especially valuable for subjective tasks like discourse analysis where boundaries can be ambiguous). We calculate the performance for each conversation and aggregate the results together to report the mean and standard deviation across all the conversations in our dataset.

\paragraph{Threading Sub-Category analysis}

Recognizing that aggregate performance metrics can mask critical weaknesses, we extend our evaluation with a detailed error analysis to pinpoint where LLMs are more likely to struggle. As described in Section \ref{threading_human_annotation}, one of our human annotators further labeled a subset of the transcripts with threading subcategories that specify the type of discourse relationship each utterance represents. This additional layer of coding enables us to analyze LLM performance conditional on the specific nature of the linkage between speaker utterances. The threading subcategories are:

\begin{itemize}
\setlength{\itemsep}{-0.2em}
    \item \textbf{AP} (\textit{Adjacency Pairs}): Simple question-answer or offer-response sequences \citep{tsui1989beyond}.
    \item \textbf{E} (\textit{Explicit Coherence Relations}): Logical connectors explicitly present in the language (e.g., \textit{“because”, “but”, “therefore”}) \citep{taboada2009implicit}.
    \item \textbf{I} (\textit{Implicit Coherence Relations}): Coherence inferred without explicit signals \citep{taboada2009implicit}.
    \item \textbf{TT} (\textit{Topic Transitions}): Utterances that signal a clear shift to a new topic or subtask \citep{riou2015methodology}.
    \item \textbf{CI} (\textit{Consensus Information}): Statements summarizing or confirming shared group understanding \citep{wooffitt2005conversation}
    \item \textbf{BC} (\textit{Backchannel Responses}): Minimal acknowledgments (e.g., \textit{“yeah”, “uh-huh”}) that often serve as ephemeral interactional cues \citep{li2010backchannel}
    \item \textbf{SC} (\textit{Self Continuation}): Cases where a speaker resumes their own train of thought after a pause or interruption.
\end{itemize}
These subcategories were developed through a combination of prior research guidance and practical refinements by our human annotators. While most subcategories drew directly from established literature and our team’s threading codebook, \textit{self-continuation} was added to capture frequent instances where speakers expand on their own earlier statements—an interaction pattern that appeared frequently in our data but was not adequately captured by the other subcategories. By analyzing performance across these nuanced subcategories, we surface systematic strengths and weaknesses in how LLMs handle different discourse structures.

\subsection{Downstream Conversational Analysis Evaluation}
\label{ABCDE_analysis}

First, we assess the overall performance of our different LLM approaches. For this, we directly compare the LLM-generated outputs from our different methods against the human-annotated ground truth labels using three complementary metrics: \textbf{accuracy}, \textbf{macro-average F1-score}, and \textbf{Cohen’s kappa}. We compute performance metrics for each conversation for category E \textit{(Elicit)} from the ABCDE framework.
Again, we focus on E alone for our analysis, given its acceptable human–human agreement, theoretical significance for collaborative learning, and broad applicability across discourse frameworks.

\paragraph{Comparison to SOTA approaches in the literature}
\begin{table}
\caption{Baselines for ABCDE experiments from recent works in the literature}
\footnotesize
\renewcommand{\arraystretch}{1.2}
\centering

\begin{tabular}{|>{\raggedright\arraybackslash}p{0.15\linewidth}|>{\raggedright\arraybackslash}p{0.75\linewidth}|}\hline
 \textbf{Paper}&\textbf{Details of their approach}\\\hline
\hline
\cite{lee2025capturing} & \textbf{Lee et al. (2025)} introduce \textit{CoComTag}, an LLM-powered annotation system for detecting collaborative competencies in student group discourse. The key method they contribute is a \textbf{structured base prompt} that combines dialogue context, detailed category definitions (based on Sun et al.’s generalized collaborative competency model), an explicit output format, and clear instructions. To improve the model’s coding accuracy, they test four \textbf{prompting techniques}:

\begin{itemize}
\setlength{\itemsep}{-0.2em}
    \item \textbf{Persona}: assigning the LLM the role of a teacher evaluating student collaboration
    \item \textbf{Predict Addressee}: requiring the LLM to infer who each utterance is addressing before coding
    \item \textbf{Chain-of-Thought}: prompting the LLM to reason about speaker intention step by step
    \item \textbf{Example Cases}: providing micro- and macro-level examples for each category as few-shot guidance
\end{itemize}
Their experiments show that this multi-component prompt design, with carefully chosen examples, improves performance compared to a simple base prompt, especially for fine-grained collaborative process coding.

To compare our results on ABCDE LLM performance, we use their \textbf{prompt structure design} as one of the baseline references. \\
\hline
\cite{qamarllms} & \textbf{Qamar et al. (2025)} investigates why LLMs struggle with fine-grained dialogue act (DA) classification in multi-party settings, proposing three \textbf{pre-tasks} — \textbf{Turn Management} (predicting who will speak next), \textbf{Communicative Function Identification} (inferring the social or pragmatic role of each utterance, such as question, suggestion, or agreement), and \textbf{Dialogue Structure Prediction} (mapping how utterances connect across turns) — as essential subtasks for improving DA accuracy. 
\newline
For our comparison, we adapt their \textbf{instruction-based prompting approach}: they provide a clear task description, detailed definitions for each DA label, and in-context examples showing how to handle surrounding speaker context. The prompt explicitly prepends speaker names and includes both preceding and (optionally) future utterances to help the model infer speaker roles and discourse structure. 
\newline
This structure helps frame DA classification as a communicative function prediction, which we contrast with our ABCDE framework. \\
\hline
\cite{martinenghi2024llms} & \textbf{Martinenghi et al., (2024)} investigates zero- and few-shot learning approaches for classifying and predicting dialogue acts (DAs) in multi-party spoken interactions. Specifically, the authors aim to explore how example-based task formulation and pragmatic features impact the performance of large language models when labeling or forecasting DAs. 
\newline
Their prompt approach provides the LLM with a short excerpt of preceding dialogue, detailed definitions for each DA category, and in-context examples that show how each act appears in conversation. The prompt also emphasizes instructions for interpreting the speaker’s intention and interaction with other participants, highlighting how local context shapes discourse roles. 
\newline
We include this example-based classification prompt as one of our baselines for comparison when evaluating how different prompting methods support LLM performance on our ABCDE coding framework. \\
\hline

\end{tabular}
\label{tab:baselines}
\end{table}

To contextualize our approach within the literature, we note that \emph{no prior work has directly examined threading as a modeling component} for multiparty conversation analysis, leaving us without established baselines for this task.
In contrast, related work exists for \textit{downstream} collaborative discourse coding, where recent state-of-the-art (SOTA) methods have leveraged LLMs for tasks such as ABCDE coding (see Table~\ref{tab:baselines}).
Because threading itself cannot be directly benchmarked against prior methods, we include downstream conversational analysis as a second evaluation setting allowing us to \textit{systematically test} whether adding thread information yields measurable performance gains.
We adapt and repurpose prompting strategies from these works to serve as comparative baselines for the E \textit{(Elicit)} \citep{lin2025abcde} conversational analysis, focusing on two common paradigms: the \textbf{All-At-Once} approach \citep{martinenghi2024llms} and the \textbf{Sliding Window} approach \citep{lee2025capturing,qamarllms}.
This dual evaluation lets us assess whether the benefits of threading hold regardless of whether the model processes dialogue in a global (All-At-Once) or local (Sliding-Window) manner.


\paragraph{Practical Trade-Offs: Time Cost Analysis}

While recent studies have shown that sliding window prompting can outperform full context, these gains often come at a cost in inference time and token usage, an important trade-off. We conduct further analysis to quantify the practical benefits and limitations of incorporating the different LLM-based coding methods into real-world qualitative research workflows and classroom formative assessments for student groups, highlighting when automated threading and coding are most feasible or when additional human oversight remains necessary.

\section{Results}
In this section, we present the LLM results for threading and conversational analysis. 

\subsection{LLM Threading Results}

\subsubsection{LLM Threading Performance}
\paragraph{Zero- and Few-Shot Learning for All Models}
We first evaluated how well each model performs the threading task when prompted in a zero-shot, one-shot, or three-shot learning setup. To remind the reader, in these conditions, the LLM receives the entire conversation transcript all at once, along with the full threading definition and codebook. In the few-shot settings, additional fully labeled transcripts are included to help the model learn recurring patterns in how threads unfold.
As described in Section \ref{threading_methods}, we ran our initial experiments in Tables \ref{tab:zero-threading} and \ref{tab:few-threading} with all OpenAI models to establish whether the reasoning models' performance trends uncovered in prior work hold true within the educational domain. 

\textbf{Performance Against Human Ground Truth. }Overall, we find that threading performance remains low across all models in the zero-shot condition, with mean accuracies ranging from 0.10–0.40 and macro-average F1-scores typically well below 0.35. Cohen’s kappa values, which adjust for chance agreement, follow similar trends and reinforce that models have difficulty reliably matching human-annotated thread labels when asked to do so end-to-end (Table \ref{tab:zero-threading}).

\textbf{Effect of Few-Shot Examples. }Providing one or three few-shot example transcripts yields only marginal improvements for most models.  This suggests that simply adding more example transcripts does not improve a model’s ability to infer thread structures in synchronous talk (Table \ref{tab:few-threading}).

\textbf{Long Transcript Lengths and Context Limits. }One likely factor limiting performance is prompt length. In this experiment, entire transcripts were provided in a single prompt, often exceeding several hundred utterances. Recent studies show that while modern LLMs can accept long inputs in theory, their effective reasoning often breaks down as context length grows. \cite{hsieh2024ruler} and \cite{liu2023lost} find that models tend to underuse information buried in the middle of long prompts, while \cite{roberts2024needle} show that maintaining coherent threads across very large contexts remains challenging. This helps explain why providing entire transcripts at once can actually reduce threading accuracy in practice. This highlights the challenge of asking LLMs to infer complex multi-party discourse structure from long transcripts all at once without incremental anchoring.

\begin{table}[H]
\centering
\caption{All-At-Once Zero shot Threading Results (mean ± standard deviation)}
\begin{tabular}{| l | l | l | l |}
\hline
\textbf{Model} & \textbf{Accuracy} & \textbf{F1 (Macro)} & \textbf{Cohen’s Kappa} \\
\hline
o3-mini & 0.1064 ± 0.0575 & 0.0478 ± 0.0304 & 0.0941 ± 0.0529 \\
\hline
o1-mini & 0.1191 ± 0.0375 & 0.0639 ± 0.0221 & 0.1066 ± 0.0338 \\
\hline
gpt-4.1 & 0.1079 ± 0.0238 & 0.0630 ± 0.0174 & 0.0998 ± 0.0239 \\
\hline
gpt-4o & 0.1093 ± 0.0276 & 0.0570 ± 0.0228 & 0.0983 ± 0.0285 \\
\hline
gpt-4.1-mini & 0.1023 ± 0.0404 & 0.0372 ± 0.0237 & 0.0830 ± 0.0328 \\
\hline
gpt-4.1-nano & 0.3112 ± 0.2375 & 0.2538 ± 0.2395 & 0.2786 ± 0.2609 \\
\hline
\textbf{gpt-4o-mini} & \textbf{0.4090 ± 0.2080} & \textbf{0.3451 ± 0.2093} & \textbf{0.4033 ± 0.2114} \\
\hline

\end{tabular}
\label{tab:zero-threading}




\centering
\caption{All-At-Once Three shot Threading Results, (mean ± standard deviation)}

\begin{tabular}{| l | l | l | l |}
\hline
\textbf{Model} & \textbf{Accuracy} & \textbf{F1 (Macro)} & \textbf{Cohen’s Kappa} \\
\hline
o3-mini & 0.1170 ± 0.1677 & 0.0667 ± 0.1501 & 0.1051 ± 0.1679 \\
\hline
o1-mini & 0.0626 ± 0.0560 & 0.0287 ± 0.0325 & 0.0516 ± 0.0547 \\
\hline
gpt-4.1 & 0.1689 ± 0.1504 & 0.1177 ± 0.1275 & 0.1621 ± 0.1521 \\
\hline
gpt-4o & 0.3016 ± 0.1364 & 0.2279 ± 0.1411 & 0.2797 ± 0.1453 \\
\hline
gpt-4.1-mini & 0.1059 ± 0.0888 & 0.0617 ± 0.0717 & 0.0895 ± 0.0869 \\
\hline
gpt-4.1-nano & 0.1212 ± 0.0601 & 0.0219 ± 0.0209 & 0.0231 ± 0.0264 \\
\hline
\textbf{gpt-4o-mini} & \textbf{0.4343 ± 0.2671} & \textbf{0.3862 ± 0.2561} & \textbf{0.4275 ± 0.2701} \\
\hline

\end{tabular}
\label{tab:few-threading}

\end{table}

\paragraph{Sliding Windows for All Models}

\begin{table}[H]



\caption{Threading Results with Sliding Window size N = 20, (mean ± standard deviation)}
\centering

\begin{tabular}{| l | l | l | l |}
\hline
\textbf{Model} & \textbf{Accuracy} & \textbf{F1 (Macro)} & \textbf{Cohen’s Kappa} \\
\hline
o3-mini & 0.6289 ± 0.1690 & 0.5434 ± 0.1587 & 0.6214 ± 0.1725 \\
\hline
o1-mini & 0.5713 ± 0.1555 & 0.4746 ± 0.1451 & 0.5601 ± 0.1606 \\
\hline
\textbf{gpt-4.1} & \textbf{0.6582 ± 0.1889} & \textbf{0.6061 ± 0.1859} & \textbf{0.6523 ± 0.1925} \\
\hline
gpt-4o & 0.6173 ± 0.1791 & 0.5517 ± 0.1740 & 0.6103 ± 0.1838 \\
\hline
gpt-4.1-mini & 0.5793 ± 0.1689 & 0.5172 ± 0.1611 & 0.5733 ± 0.1721 \\
\hline
gpt-4.1-nano & 0.5479 ± 0.1643 & 0.4896 ± 0.1501 & 0.5429 ± 0.1659 \\
\hline
gpt-4o-mini & 0.4250 ± 0.1266 & 0.3389 ± 0.1152 & 0.4146 ± 0.1335 \\
\hline

\end{tabular}
\label{tab:threading-sliding}

\end{table}

Next, we introduced sliding window based prompting strategies — \textbf{sliding window (N=20)}  to assess whether sliding windows can help LLMs improve threading accuracy over end-to-end zero- or few-shot prompting. Overall, the inclusion of sliding windows  produced significantly higher performance than the base conditions: across models, mean accuracies rose to $\sim$0.50–0.65, macro F1-scores improved to$\sim$0.50–0.60, and Cohen’s kappa values approached or exceeded 0.60 for the best models (Table \ref{tab:threading-sliding}).

\texttt{GPT-4.1} showed the strongest results, achieving a kappa of \textbf{0.6523}. Other models like o3-mini also benefited, with accuracy jumping to 62\% and kappa \~0.62, despite underperforming in the zero-shot setting.

\textbf{These results suggests that providing explicit linkage history within a window \textit{may} help LLMs maintain more stable discourse reasoning over multiple turns — aligning with how human coders use local context and recent thread decisions to disambiguate overlapping contributions.}

These findings reinforce the idea that breaking long synchronous transcripts into manageable context windows — and supplying explicit thread signals where possible — significantly boosts performance compared to attempting the task in a single pass. Yet they also suggest that even with these improvements, threading remains a challenging discourse task for LLMs in natural, multi-party conversation — motivating the need for continued refinement of model-assisted approaches.

\paragraph{Varying Sliding Window Sizes on Best Performing Model}

To better understand how context length influences threading accuracy, we compared three sliding window sizes (\textbf{10}, \textbf{20}, and \textbf{30} utterances) on the best-performing model, \textbf{gpt-4.1.} We found that all three window sizes yielded relatively stable and similar results, with Cohen’s kappa consistently ranging between $\sim$0.65–0.67. The performance was highest with a window size of 10 (\textbf{0.6746} kappa) and dipped slightly as the window increased to 20 and 30 (Table \ref{tab:threading-sliding_vary}). 

These results suggest that, for this context of small-group transcripts, increasing the context window did not yield substantial gains for threading accuracy. Future work should test whether similar window-size effects hold across other synchronous conversational corpora.

\begin{table}[H]

\caption{Threading results with different window sizes, (mean ± standard deviation)}
\centering

\begin{tabular}{| l | l | l | l | l |}
\hline
\textbf{Window Size} & \textbf{Model} & \textbf{Accuracy} & \textbf{F1 (Macro)} & \textbf{Cohen’s Kappa} \\
\hline
\best{\textbf{10}} & \best{\textbf{gpt-4.1}} & \best{\textbf{0.6808 ± 0.1649}} & \best{\textbf{0.6190 ± 0.1607}} & \best{\textbf{0.6746 ± 0.1684}} \\
\hline
20 & gpt-4.1 & 0.6582 ± 0.1889 & 0.6061 ± 0.1859 & 0.6523 ± 0.1925 \\
\hline
30 & gpt-4.1 & 0.6722 ± 0.1666 & 0.6171 ± 0.1615 & 0.6669 ± 0.1696 \\
\hline

\end{tabular}
\label{tab:threading-sliding_vary}

\end{table}

\subsubsection{LLM Threading Error Analysis}

As described in Section \ref{threading_analysis}, one of our human coders added exploratory sub-category labels to a subset of transcripts to unpack where LLMs struggle most with threading in small-group synchronous talk. This additional coding was performed on three transcripts spanning both student groups in the data science workshop dataset.

Across all tested models (Table \ref{tab:threading_error_table_all} in appendix), some thread types proved consistently easier to detect than others. We present the results of the best performing model \textit{gpt-4.1} here in Table \ref{tab:threading_error_table_4_1}.
LLM performance was the highest on utterances involving 
\textbf{Adjacency pairs (AP)}, \textbf{explicit coherence relations (E)}, and\textbf{ Backchannel responses (BC)}. These showed the highest agreement with human annotations (with kappas $\geq$ 0.90). On the other hand, \textbf{Topic transitions (TT)} and \textbf{Self-continuation (SC)} responses were the hardest for LLMs to reliably thread with kappas of 0.25 and 0.43 respectively. We unpack these results further in the discussion. 

\begin{table}
\centering
\caption{Threading Error Analysis: Sub-category Performance for \texttt{GPT-4.1}, (mean ± standard deviation)}
\begin{tabular}{|l|c|c|c|}
\hline
\textbf{Category} & \textbf{Accuracy} & \textbf{F1 (Macro)} & \textbf{Cohen's Kappa} \\
\hline

AP & \textbf{0.9065} $\pm$ 0.0535 & \textbf{0.8709} $\pm$ 0.0522 & \textbf{0.9034} $\pm$ 0.0550 \\\hline
E  & \textbf{0.9444} $\pm$ 0.0556 & \textbf{0.9000} $\pm$ 0.1000 & \textbf{0.9384} $\pm$ 0.0616 \\\hline
I  & 0.8280 $\pm$ 0.0087 & 0.7454 $\pm$ 0.0087 & 0.8261 $\pm$ 0.0090 \\\hline
TT & 0.3997 $\pm$ 0.1388 & 0.1782 $\pm$ 0.1551 & 0.2567 $\pm$ 0.2127 \\\hline
CI & 0.7500 $\pm$ 0.0000 & 0.6000 $\pm$ 0.0000 & 0.6923 $\pm$ 0.0000 \\\hline
BC & \textbf{0.9615} $\pm$ 0.0385 & \textbf{0.9444} $\pm$ 0.0556 & \textbf{0.9578} $\pm$ 0.0422 \\\hline
SC & 0.4596 $\pm$ 0.0960 & 0.3194 $\pm$ 0.0972 & 0.4309 $\pm$ 0.0891 \\ \hline

\end{tabular}
\label{tab:threading_error_table_4_1}
\end{table}

\begin{table}[H]
\centering
\caption{Comparison of LLM Performance on E (\textit{Elicit}) Annotation: Mean ± Standard Deviation for Accuracy, F1 (Macro), and Cohen’s Kappa across Approaches }
\begin{tabular}{|l|c|c|c|}
\hline
\textbf{Approach} & \textbf{Accuracy} & \textbf{F1} & \textbf{Kappa} \\
\hline
No Threads (All-at-Once) & 0.6761 ± 0.0611 & 0.2131 ± 0.0753 & 0.0502 ± 0.0408 \\
\hline
LLM Threads (All-at-Once) & 0.8524 ± 0.0570 & 0.7360 ± 0.1062 & 0.6325 ± 0.1420 \\
\hline
Human Threads (All-at-Once) & 0.8486 ± 0.0619 & 0.7240 ± 0.1351 & 0.6192 ± 0.1717 \\
\hline
No Threads (Sliding Window) & 0.8341 ± 0.0566 & 0.7169 ± 0.1177 & 0.6000 ± 0.1515 \\
\hline
\best{\textbf{LLM Threads (Sliding Window)}} & \best{\textbf{0.8546 ± 0.0553}} & \best{\textbf{0.7409 ± 0.1140}} & \best{\textbf{0.6395 ± 0.1479}} \\
\hline
Human Threads (Sliding Window) & 0.8499 ± 0.0487 & 0.7280 ± 0.1117 & 0.6239 ± 0.1391 \\
\hline
Baseline: Lee et al. & 0.6573 ± 0.0483 & 0.1922 ± 0.0868 & 0.0024 ± 0.0571 \\
\hline
Baseline: Qamar et al. & 0.7910 ± 0.0886 & 0.6529 ± 0.1565 & 0.5037 ± 0.2144 \\
\hline
Baseline: Martinegni et al. & 0.6570 ± 0.0462 & 0.3211 ± 0.0879 & 0.0989 ± 0.0879 \\
\hline
\end{tabular}
\label{tab:LLM_E_ABCDE}
\end{table}

\subsection{LLM Conversational Analysis Results}
\subsubsection{LLM Conversational Analysis Performance}

The full set of results for E (\textit{Elicit}) annotation experiments with LLMs are shown in Table \ref{tab:LLM_E_ABCDE}. 

Across all experimental settings, incorporating explicit conversational threads consistently enhances performance. In the All-at-once setups, \textbf{adding thread labels boosts Cohen’s kappa by a whopping 58\%} compared to the no-threads baseline. For the Sliding Window approaches, we also observe performance gains on average—segmenting conversations into smaller windows allows the LLM to better preserve local context and follow relational cues, which is particularly valuable for lengthy, multi-party transcripts. Among these, the Sliding Window condition with LLM threads achieves the strongest results, with a mean F1 of 0.74 and Cohen’s kappa of 0.63. These findings indicate that high-quality thread annotations can further improve downstream coding accuracy, underscoring that detecting discourse moves from frameworks like ABCDE are heavily reliant on precise conversational linkage.

When compared to prior baselines adapted from \citep{lee2025capturing,qamarllms,martinenghi2024llms}, all of our threading-enhanced approaches also consistently outperform these benchmarks on average. This further underscores that simple sequence classification models without access to explicit conversational linkage information struggle to capture the relational nature of collaborative discourse.

\begin{figure}[htb]
  \centering
  \includegraphics[width=1\textwidth]{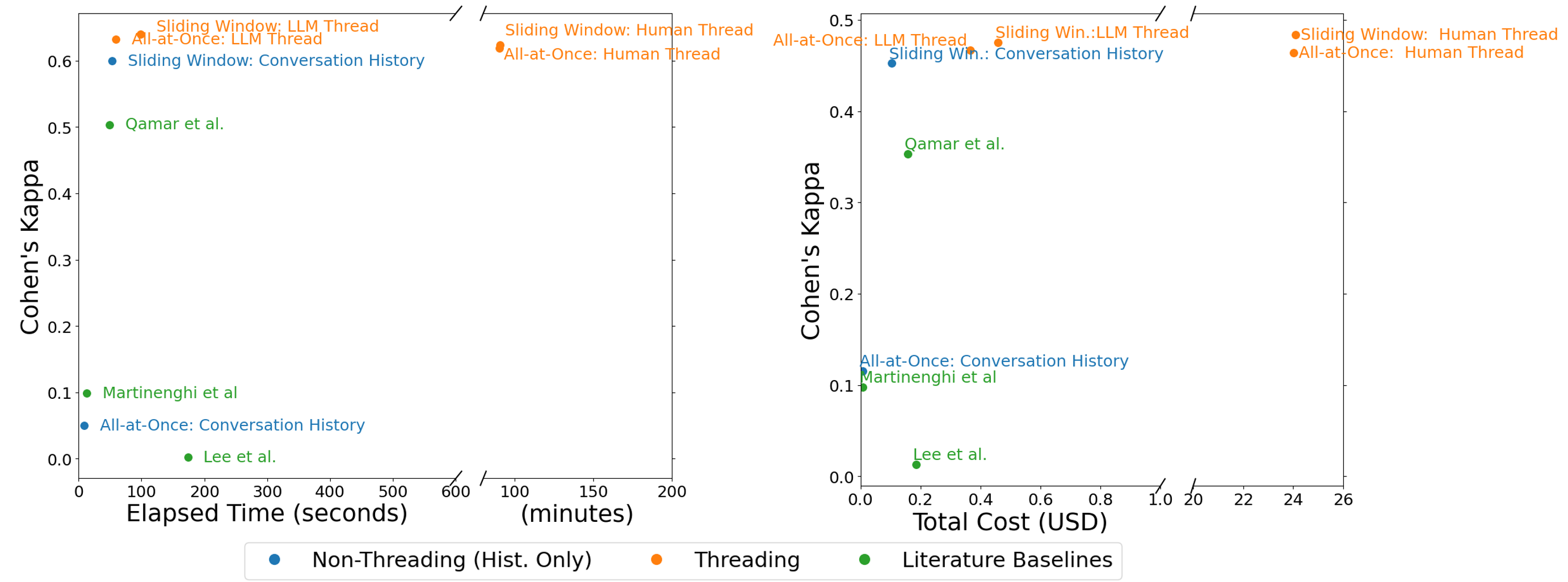}
  \caption{Time/Cost vs. Performance Tradeoff for average of categories with best interrator reliability \emph{E: Eliciting responses or actions}. Performance: Left: Time vs. Cohen's Kappa for  Right: Monetary Cost vs. Cohen's Kappa}
  \label{fig:cost}
\end{figure}

\subsubsection{Conversational Analysis Performance vs Time \& Cost}
\label{abcde_cost}

Figure \ref{fig:cost} compares Cohen’s Kappa scores for the E (\textit{Eliciting responses or actions}) code against two practical constraints: inference time (left) and monetary cost (right). Both sliding window configurations—whether using human-generated threads or LLM-generated threads—consistently yield the highest performance, reaching Cohen’s Kappa $\approx 0.60$, well above prior literature baselines (shown in green), which generally fall below 0.35.

While human-threaded sliding window runs nearly match LLM-threaded runs in accuracy, they are dramatically more resource-intensive: annotating a one-hour transcript takes human coders at least 1.5 hours, translating to roughly \$25 USD at U.S. minimum wage rates. In contrast, LLM-threaded approaches complete the same task in seconds to a couple of minutes and at a fraction of the cost (under \$1 per transcript in our setting). 

We will now discuss the implications of these results in relation to prior work.

\section{Discussion}
This section presents a deeper examination of the results from the LLM-generated threading and conversational analysis annotations, and considers their implications for the design and application of such models in the future.

\subsection{LLM Threading: Strengths and Areas for Improvement}

Our results show that LLMs are able to reach acceptable inter-rator agreement against the human ground-truth thread labels using our threading guidebook. However, performance is not consistent across the different thread categories. 
\textbf{Adjacency pairs (AP)} \citep{tsui1989beyond} and \textbf{explicit coherence relations (E)} \citep{taboada2009implicit} showed the highest agreement with human annotations. This is unsurprising since these discourse moves are often marked by clear question–response structures or explicit conjunctions that LLMs can recognize directly in text.

\textbf{Consensus information (CI)} \citep{wooffitt2005conversation} and \textbf{backchannel responses (BC)} \citep{li2010backchannel} also showed relatively strong performance, since both often appear in predictable, formulaic phrasing. However, our human coders noted an important subtlety: while \textbf{backchannels themselves} (e.g., “yeah,” “right,” “uh-huh”) are easy for models to detect, correctly threading the \textit{responses that follow these interruptions} — especially when speakers resume a prior idea — is much harder. This is where \textbf{self-continuation (SC)} threads often break down. For example, if a student is interrupted by a backchannel and then picks up their thought, the LLM must infer that the new utterance should link back to the original statement, not the backchannel. This depends heavily on group dynamics: more socially connected groups with free-flowing back-and-forth tend to interleave ideas, which increases threading ambiguity. In our data involving socially inclined high school students, SC consistently showed some of the lowest kappa scores across models, highlighting this weakness.

\textbf{Implicit coherence (I)} \citep{taboada2009implicit} was moderately challenging for LLMs. These connections require the model to infer logical flow without clear linking words in the language, demanding more nuanced contextual reasoning.

\textbf{Topic transitions (TT)} \citep{riou2015methodology} were among the most error-prone categories. Human coders noted that defining when a group truly shifts topics is inherently ambiguous in less structured settings, like the exploratory discussions in the data science workshop. Sub-topic boundaries are fluid, and speakers rarely signal transitions with explicit markers — making it easy for LLMs to miss these cues.

Finally, the presence of the conversational agent \textit{Oscar }in our dataset’s conversations added additional complexity. Coders noted that when the agent lagged in its outputs (owing to technological limitations), it created discontinuities that muddled thread boundaries.  If the agent responded coherently and without delay, its presence did not noticeably interfere; but when it injected irrelevant or delayed statements, the group’s reaction often determined whether the thread could be reliably traced.

\textbf{Taken together, these patterns show that while LLMs can handle simple, well-signaled linkages, they still struggle with conversational messiness: subtle topic shifts, implicit cues, and fragmented turns spread across interruptions.}

\subsection{Unpacking LLM Downstream Conversational Analysis}

As mentioned in section \ref{ABCDE-intro}, we picked E(\textit{Elicit}) from the ABCDE framework for this paper. Its human kappa was the highest at 0.881 (table \ref{tab:human_ABCDE}). This means human coders can clearly identify elicitation moves — questions, prompts, or invitations to speak — thanks to overt markers like interrogatives (\textit{“what,” “how,” “can you…”}) or imperative forms (\textit{“tell me,” “explain…”}).

LLM-based annotation of E achieves its best performance across all tested conditions when provided with explicit thread information. In the sliding window setup, giving access to LLM-generated thread labels reaches the highest observed agreement. Threading likely helps here because elicitation moves often occur in response to—or in anticipation of—a specific conversational exchange. 

However, performance relative to human annotation is still far from perfect: the LLM may miss rhetorical questions \citep{kuhn2024elephant}, indirect prompts, or cases where elicitation is context-dependent rather than syntactically distinguishable \citep{medkova2020automatic}, showing room for improvement. Prior work has emphasized the need for better semantic, syntactic, and pragmatic understanding to detect rhetoric figures \citep{ranganath2018understanding}. These cases highlight that even for a relatively well-marked code like E, threading provides crucial scaffolding for interpretation — but richer discourse understanding is still needed to approach human-level agreement.

\paragraph{Implications of Performance vs. Time \& Cost}

 Consistent with prior findings that sliding window prompting can outperform full-context approaches \citep{tran2024multi}, our results show that both human-threaded and LLM-threaded sliding window methods achieve the highest Kappa scores among the automated conditions. 
However, particularly for human-threaded runs, the performance comes with a steep increase in both elapsed time and total cost, as it requires substantial and costly manual annotations in addition to model inference.

In contrast, the LLM-generated threading then conducting conversational analysis approach offers a more balanced trade-off, consistently improving over non-threaded conversational history approaches, while keeping time and cost considerably lower than human-treaded pipelines. This makes it a more effective option for large-scale datasets or frequent classroom assessments, where resources are limited.

\subsection{Implications for Future Work}

Our results show that stronger performance on widely used general-purpose LLM benchmarks does not necessarily translate into better outcomes for complex conversational analysis tasks. For example, reasoning-oriented models such as \textbf{o3-mini}, despite excelling in code and math reasoning \citep{hendrycksmeasuring,jimenez2023swe}, did not outperform \textbf{GPT-4.1} on our threading and ABCDE coding tasks. This underscores that alignment to the specific domain and task—here, educational discourse and conversational analysis—matters as much as, if not more than, general reasoning ability. Evaluating models directly on domain-specific datasets and metrics, as we have done, is therefore essential before assuming performance will generalize from loosely related benchmarks.

In this paper, thread annotations—whether human- or LLM-generated—were provided to the LLMs for ABCDE annotations as static, pre-computed inputs. A promising future extension would be a chained, two-step inference process in which the same LLM first generates thread structures for a conversation and then, within the same session and without clearing context, codes the final utterance for the ABCDEs using its own threading output. This approach could allow the model to leverage its fresh, self-generated discourse segmentation as in-context reasoning, potentially improving performance.  Even in a single-step setting, encouraging \textbf{implicit threading}—prompting the model to reason through conversational structure before producing a code (e.g. \textit{``You should thread the conversation before giving me a ABCDE label"}) —could mimic the process used by human annotators. Advances in \textbf{frontier long-context model architectures} \citep{hsieh2024ruler,liu2023lost,roberts2024needle,zhang2024bench} also warrant investigation, as they may reduce or eliminate the need for sliding windows altogether.

At the sub-task level, certain thread types—such as \textbf{self-continuations} or \textbf{topic transitions}—remain particularly challenging for LLMs. Future work could explore targeted sub-prompts or fine-tuning to better capture these discourse patterns. While this study concentrated on the \textbf{E (\textit{Elicit})} code as a clear well-bounded pilot investigation of how threading impacts automated coding for a socially and pedagogically critical discourse move, it also lays the groundwork for extending our methodology to the full ABCDE and other conversational analysis frameworks in the future. 

Finally, while our evaluation focused on models within a single LLM family (namely, OpenAI) to minimize pretraining and system-level confounds, expanding this analysis to include diverse LLM providers and open-weight models remains an important direction for future work.

\section{Conclusion}
This work advances the study of collaborative discourse by operationalizing conversational threads as a unit of analysis for synchronous small-group dialogue. We propose a systematic guidebook for thread annotation, demonstrate that large language models (LLMs) can partially recover these structures using sliding-window prompting, and show that threading, especially when accurate, substantially improves downstream deductive coding performance.

Our results reveal that while threading remains a difficult task for LLMs in natural, multi-party contexts, providing even imperfect thread labels enhances model reliability on relational discourse frameworks like ABCDE. This suggests that explicit discourse structure serves as a critical scaffold for LLM-based collaborative learning analysis.

Practically, we find that using LLM-generated threads offers a strong balance of accuracy, cost-efficiency, and scalability, making it a viable tool for researchers, educators, and developers aiming to support collaborative sense-making in real time or at scale. Future work should explore how thread quality interacts with other discourse frameworks and whether hybrid approaches, combining automated threading with targeted human correction, can further optimize performance.

\bibliographystyle{apacite}
\bibliography{ref}

\section{Appendices}




\subsection{Threading Prompt:} 
\label{app:prompts}

You are an expert qualitative coder and your job is to label threads within a transcript from an in-person conversation. I will provide you with a transcript segment (with some previous thread annotations) and your task is to assign the appropriate \texttt{respond\_line} value to the \textbf{last utterance only}.

\vspace{0.5em}
\textbf{Definition:} \\
A conversation thread is a sequence of connected messages or utterances produced in relation to the same theme or topic. It is sequentially organized across interactants’ turns, with each turn contributing to the thread. A turn is one uninterrupted utterance, and a contribution is the portion of a turn that belongs to a single thread. Therefore, a turn can contain one or more contributions.

\vspace{0.5em}
\textbf{Instructions:}
\begin{itemize}
  \item Take the first contribution as the beginning of the first thread.
  \item If the next contribution is a reply or response to the first contribution, then it is a continuation of the first thread.
  \item A reply is a contribution that answers a question or clarifies a statement or issue.
  \item If a contribution is neither on the same topic nor a reply or response to an earlier contribution, then it is a new thread.
  \item As a ground rule for threading, always start with the immediate previous utterance and work your way up, eliminating utterances one after the other until you find one that works as a link to the current line.
\end{itemize}

\vspace{0.5em}
\textbf{Edge Cases from the Threading Codebook:}
\begin{enumerate}
  \item \textbf{Backchannel responses:} 1-2 word acknowledgements (e.g., “yeah”, “hmmm”) should be skipped. Link back to the last substantial utterance.
  
  \textit{Example:}
  \begin{quote}
  \#1 Prerna: As I was saying we need to continue working on this for the rest of this week… \texttt{[respond\_line= -]} \\
  \#2 Carúmey: yeah \texttt{[respond\_line= 1]} \\
  \#3 Prerna: We need to make progress on the following topics… \texttt{[respond\_line= 1]}
  \end{quote}
  
  \item \textbf{Transition statements} should start new threads if they introduce new topics (e.g., "Now, let’s move on...") — label with \texttt{-}.
  
  \item \textbf{Split threads in one utterance:} Use two labels separated by a comma. E.g., \texttt{(45, -)}.
  
  \item \textbf{Consensus statements:} These summarize decisions or clarify. Link to the last referred utterance.
      
  \textit{Example:}
  \begin{quote}
  \#23 Red Morgan: ...? \texttt{[respond\_line= 22]} \\
  \#24 Alex: Okay, I am fine with that. \texttt{[respond\_line= 23]} \\
  \#25 Red Morgan: Our current order is... What do you think? \texttt{[respond\_line= (24, -)]}
  \end{quote}
  
  \item Use timestamps to resolve overlapping responses to the same utterance.
  
  \item Link self-directed or rambling speech through interruptions using the same thread.
  
  \item If two equally valid options exist, prefer the closer (lower distance).
  
  \item Addressing someone does not mean replying to their last utterance—consider topical relevance.
\end{enumerate}

\vspace{0.5em}
Please review the definition, instructions, and areas to keep in mind. Then I will provide the example transcript with labels and new transcript without labels for threading.

\vspace{1em}
\texttt{<<<TRANSCRIPT\_START>>>} \\
\texttt{\{Sliding Window Transcript\}} \\
\texttt{<<<TRANSCRIPT\_END>>>}

\vspace{0.5em}
Label \textbf{ONLY the last utterance} in this format: \\
\texttt{\{Utterance Number\} \{Speaker Name\} [respond line = N]} \\
Where \texttt{N} is the line number it responds to (e.g., 10), or \texttt{-} if it starts a new thread. Use \texttt{(X, -)} if there are two parts. Do not explain or include anything else.

\subsection{Threading Error Results- All Models}

\begin{table}[H]
\tiny
\centering
\caption{Threading Error Analysis: Sub-category Performance by Model, (mean ± standard deviation)}
\renewcommand{\arraystretch}{1.3}
\begin{tabular}{|l|c|c|c|}
\hline
\textbf{Category} & \textbf{Accuracy} & \textbf{F1 (Macro)} & \textbf{Cohen's Kappa} \\
\hline
 \multicolumn{4}{|c|}{o3-mini}\\
\hline

AP & 0.8024 $\pm$ 0.0376 & 0.6998 $\pm$ 0.0243 & 0.7967 $\pm$ 0.0378 \\\hline
E  & \textbf{0.8333} $\pm$ 0.1667 & \textbf{0.7500} $\pm$ 0.2500 & \textbf{0.8200} $\pm$ 0.1800 \\\hline
I  & 0.7311 $\pm$ 0.0444 & 0.6186 $\pm$ 0.0512 & 0.7284 $\pm$ 0.0450 \\\hline
TT & 0.4264 $\pm$ 0.0351 & 0.0853 $\pm$ 0.0508 & 0.1776 $\pm$ 0.1002 \\\hline
CI & 0.7500 $\pm$ 0.0000 & 0.6000 $\pm$ 0.0000 & 0.6923 $\pm$ 0.0000 \\\hline
BC & \textbf{1.0000} $\pm$ 0.0000 & \textbf{1.0000} $\pm$ 0.0000 & \textbf{1.0000} $\pm$ 0.0000 \\\hline
SC & 0.4596 $\pm$ 0.0960 & 0.3034 $\pm$ 0.0812 & 0.4341 $\pm$ 0.0922 \\
\hline
 \multicolumn{4}{|c|}{o1-mini}\\
\hline

AP & 0.8276 $\pm$ 0.0924 & 0.7435 $\pm$ 0.1084 & 0.8225 $\pm$ 0.0945 \\\hline
E  & 0.7083 $\pm$ 0.0417 & 0.5714 $\pm$ 0.0714 & 0.6857 $\pm$ 0.0457 \\\hline
I  & 0.6291 $\pm$ 0.0749 & 0.5118 $\pm$ 0.0885 & 0.6255 $\pm$ 0.0759 \\\hline
TT & 0.4431 $\pm$ 0.0953 & 0.1248 $\pm$ 0.0931 & 0.2348 $\pm$ 0.1698 \\\hline
CI & 0.7500 $\pm$ 0.0000 & 0.6000 $\pm$ 0.0000 & 0.6923 $\pm$ 0.0000 \\\hline
BC & \textbf{0.9282} $\pm$ 0.0051 & \textbf{0.8500} $\pm$ 0.0167 & \textbf{0.9216} $\pm$ 0.0066 \\\hline
SC & 0.5152 $\pm$ 0.1515 & 0.3838 $\pm$ 0.1616 & 0.4885 $\pm$ 0.1466 \\
\hline
 \multicolumn{4}{|c|}{gpt-4.1}\\
\hline

AP & \textbf{0.9065} $\pm$ 0.0535 & \textbf{0.8709} $\pm$ 0.0522 & \textbf{0.9034} $\pm$ 0.0550 \\\hline
E  & \textbf{0.9444} $\pm$ 0.0556 & \textbf{0.9000} $\pm$ 0.1000 & \textbf{0.9384} $\pm$ 0.0616 \\\hline
I  & \textbf{0.8280} $\pm$ 0.0087 & \textbf{0.7454} $\pm$ 0.0087 & \textbf{0.8261} $\pm$ 0.0090 \\\hline
TT & 0.3997 $\pm$ 0.1388 & 0.1782 $\pm$ 0.1551 & 0.2567 $\pm$ 0.2127 \\\hline
CI & 0.7500 $\pm$ 0.0000 & 0.6000 $\pm$ 0.0000 & 0.6923 $\pm$ 0.0000 \\\hline
BC & \textbf{0.9615} $\pm$ 0.0385 & \textbf{0.9444} $\pm$ 0.0556 & \textbf{0.9578} $\pm$ 0.0422 \\\hline
SC & 0.4596 $\pm$ 0.0960 & 0.3194 $\pm$ 0.0972 & 0.4309 $\pm$ 0.0891 \\
\hline
\multicolumn{4}{|c|}{gpt-4o} \\
\hline
AP & 0.8265 $\pm$ 0.0265 & 0.7526 $\pm$ 0.0859 & 0.8209 $\pm$ 0.0275 \\\hline
E  & \textbf{0.8889} $\pm$ 0.1111 & \textbf{0.8182} $\pm$ 0.1818 & \textbf{0.8784} $\pm$ 0.1216 \\\hline
I  & \textbf{0.7700} $\pm$ 0.0251 & \textbf{0.6723} $\pm$ 0.0442 & \textbf{0.7677} $\pm$ 0.0252 \\\hline
TT & 0.3779 $\pm$ 0.1605 & 0.1758 $\pm$ 0.1576 & 0.2522 $\pm$ 0.2172 \\\hline
CI & 0.5000 $\pm$ 0.0000 & 0.3333 $\pm$ 0.0000 & 0.4286 $\pm$ 0.0000 \\\hline
BC & \textbf{0.9282} $\pm$ 0.0051 & \textbf{0.9000} $\pm$ 0.0111 & \textbf{0.9221} $\pm$ 0.0065 \\\hline
SC & 0.5051 $\pm$ 0.0505 & 0.3554 $\pm$ 0.0613 & 0.4755 $\pm$ 0.0445 \\
\hline
\multicolumn{4}{|c|}{gpt-4.1-mini} \\
\hline
AP & \textbf{0.7824} $\pm$ 0.0176 & \textbf{0.6689} $\pm$ 0.0023 & \textbf{0.7758} $\pm$ 0.0176 \\\hline
E  & 0.7083 $\pm$ 0.0417 & 0.5671 $\pm$ 0.0519 & 0.6832 $\pm$ 0.0481 \\\hline
I  & \textbf{0.7487} $\pm$ 0.0344 & \textbf{0.6450} $\pm$ 0.0463 & \textbf{0.7461} $\pm$ 0.0345 \\\hline
TT & 0.2191 $\pm$ 0.0886 & 0.1231 $\pm$ 0.1122 & 0.1550 $\pm$ 0.1359 \\\hline
CI & 0.7500 $\pm$ 0.0000 & 0.6000 $\pm$ 0.0000 & 0.6923 $\pm$ 0.0000 \\\hline
BC & \textbf{0.8564} $\pm$ 0.0103 & \textbf{0.7933} $\pm$ 0.0016 & \textbf{0.8456} $\pm$ 0.0122 \\\hline
SC & 0.3939 $\pm$ 0.0606 & 0.2542 $\pm$ 0.0399 & 0.3649 $\pm$ 0.0662 \\
\hline
\multicolumn{4}{|c|}{gpt-4.1-nano} \\
\hline
AP & \textbf{0.7624} $\pm$ 0.0024 & \textbf{0.6372} $\pm$ 0.0150 & \textbf{0.7549} $\pm$ 0.0029 \\\hline
E  & \textbf{0.8611} $\pm$ 0.0278 & \textbf{0.7571} $\pm$ 0.0429 & \textbf{0.8488} $\pm$ 0.0279 \\\hline
I  & \textbf{0.6885} $\pm$ 0.0258 & \textbf{0.5829} $\pm$ 0.0267 & \textbf{0.6854} $\pm$ 0.0263 \\\hline
TT & 0.1756 $\pm$ 0.1321 & 0.1195 $\pm$ 0.1158 & 0.1484 $\pm$ 0.1425 \\\hline
CI & 0.7500 $\pm$ 0.0000 & 0.6000 $\pm$ 0.0000 & 0.6923 $\pm$ 0.0000 \\\hline
BC & \textbf{0.8897} $\pm$ 0.0436 & \textbf{0.8530} $\pm$ 0.0581 & \textbf{0.8810} $\pm$ 0.0476 \\\hline
SC & 0.1111 $\pm$ 0.1111 & 0.0667 $\pm$ 0.0667 & 0.0962 $\pm$ 0.0962 \\
\hline
\multicolumn{4}{|c|}{gpt-4o-mini} \\
\hline
AP & \textbf{0.6982} $\pm$ 0.0218 & \textbf{0.5673} $\pm$ 0.0216 & \textbf{0.6891} $\pm$ 0.0221 \\\hline
E  & 0.5000 $\pm$ 0.1667 & 0.3690 $\pm$ 0.1548 & 0.4702 $\pm$ 0.1715 \\\hline
I  & \textbf{0.5684} $\pm$ 0.0582 & \textbf{0.4314} $\pm$ 0.0771 & \textbf{0.5641} $\pm$ 0.0582 \\\hline
TT & 0.3946 $\pm$ 0.2207 & 0.2252 $\pm$ 0.2107 & 0.2681 $\pm$ 0.2867 \\\hline
CI & 0.2500 $\pm$ 0.0000 & 0.1429 $\pm$ 0.0000 & 0.2000 $\pm$ 0.0000 \\\hline
BC & \textbf{0.5205} $\pm$ 0.2128 & \textbf{0.4074} $\pm$ 0.1852 & \textbf{0.5046} $\pm$ 0.2137 \\\hline
SC & 0.4495 $\pm$ 0.0051 & 0.2881 $\pm$ 0.0060 & 0.4195 $\pm$ 0.0116 \\
\hline

\end{tabular}
\label{tab:threading_error_table_all}
\end{table}

\subsection{ABCDE Sliding Window: Conversation Prompt} 

You are a trained qualitative researcher annotating small-group transcripts using the ABCDE Codebook. Your task is to \textbf{label only the final utterance} in the provided transcript segment using the relevant ABCDE codes below. Multiple labels may apply, or none.

\vspace{0.5em}
\textbf{Context:} \\
This is from a group of high school students working on data science tasks for analyzing data and creating visualizations in Excel/Sheets. There is also an AI agent named Oscar involved whose responses are often laggy and irrelevant.

\vspace{0.5em}
\textbf{ABCDE Codebook:}
\begin{itemize}
  \item \textbf{A (Agree):} Expresses agreement with another’s idea. Yes/no question responses don't count. Acknowledgments without clear reasoning don’t count either.
  \item \textbf{B (Build on):} Adds information to ideas proposed by someone else. A speaker building on themselves doesn’t count.
  \item \textbf{C (Chat/Comment):}
    \begin{itemize}
      \item \textbf{Chat:} Off-task talk (e.g., life, jokes)
      \item \textbf{Comment:} Reactions like ``That’s funny’’ or ``Oh, cool.''
    \end{itemize}
  \item \textbf{D (Differing Perspective):} Offers a different or opposing view.
  \item \textbf{E (Elicit Response):} Seeks info, feedback, or action (e.g., ``What do you think?’’, ``Can you\ldots?’’)
\end{itemize}

\vspace{0.5em}
To support accurate labeling, you are also provided with the conversation for the final utterance.

\vspace{1em}
\texttt{<<<TRANSCRIPT\_START>>>} \\
\texttt{\{Sliding Window Transcript\}} \\
\texttt{<<<TRANSCRIPT\_END>>>}

\vspace{0.5em}
\textbf{Label Format:}
\begin{itemize}
  \item Only label the \textbf{Target Utterance}
  \item Format: \texttt{\{line\_number\} \{speaker\} [A, B, ...]} or \texttt{[]} if no codes apply.
  \item Only one output line. Use exact line number and speaker name. No extra text.
\end{itemize}

\vspace{0.5em}
\texttt{<<<TARGET\_START>>>} \\
\texttt{ "\{Timestamp\} \{Speaker Name\} \{Utterance\}"} \\
\texttt{<<<TARGET\_END>>>}

\subsection{ABCDE - Sliding Window: Thread Prompt}

You are a trained qualitative researcher annotating small-group transcripts using the ABCDE Codebook. Your task is to \textbf{label only the final utterance} in the provided transcript segment using the relevant ABCDE codes below. Multiple labels may apply, or none.

\vspace{0.5em}
\textbf{Context:} \\
This is from a group of high school students working on data science tasks for analyzing data and creating visualizations in Excel/Sheets. There is also an AI agent named Oscar involved whose responses are often laggy and irrelevant.

\vspace{0.5em}
\textbf{ABCDE Codebook:}
\begin{itemize}
  \item \textbf{A (Agree):} Expresses agreement with another’s idea. Yes/no question responses don't count. Acknowledgments without clear reasoning don’t count either.
  \item \textbf{B (Build on):} Adds information to ideas proposed by someone else. A speaker building on themselves doesn’t count.
  \item \textbf{C (Chat/Comment):}
    \begin{itemize}
      \item \textbf{Chat:} Off-task talk (e.g., life, jokes)
      \item \textbf{Comment:} Reactions like ``That’s funny’’ or ``Oh, cool.''
    \end{itemize}
  \item \textbf{D (Differing Perspective):} Offers a different or opposing view.
  \item \textbf{E (Elicit Response):} Seeks info, feedback, or action (e.g., ``What do you think?’’, ``Can you\ldots?’’)
\end{itemize}

\vspace{0.5em}
To support accurate labeling, you are also provided with the \textbf{thread of conversation} for the final utterance.

\vspace{1em}
\texttt{<<<TRANSCRIPT\_START>>>} \\
\texttt{\{Threaded Transcript\}} \\
\texttt{<<<TRANSCRIPT\_END>>>}

\vspace{0.5em}
\textbf{Label Format:}
\begin{itemize}
  \item Only label the \textbf{Target Utterance}
  \item Format: \texttt{\{line\_number\} \{speaker\} [A, B, ...]} or \texttt{[]} if no codes apply.
  \item Only one output line. Use exact line number and speaker name. No extra text.
\end{itemize}

\vspace{0.5em}
\texttt{<<<TARGET\_START>>>} \\
\texttt{ "\{Timestamp\} \{Speaker Name\} \{Utterance\}"} \\
\texttt{<<<TARGET\_END>>>}

\subsection{ABCDE - Full Context: Conversation Prompt:}

You are a trained qualitative researcher annotating small-group transcripts using the ABCDE Codebook. Your task is to \textbf{label only the final utterance} in the provided transcript segment using the relevant ABCDE codes below. Multiple labels may apply, or none.

\vspace{0.5em}
\textbf{ABCDE Codebook:}
\begin{itemize}
  \item \textbf{A (Agree):} Expresses agreement with another’s idea. Yes/no question responses don't count. Acknowledgments without clear reasoning don’t count either.
  \item \textbf{B (Build on):} Adds information to ideas proposed by someone else. A speaker building on themselves doesn’t count.
  \item \textbf{C (Chat/Comment):}
    \begin{itemize}
      \item \textbf{Chat:} Off-task talk (e.g., life, jokes)
      \item \textbf{Comment:} Reactions like ``That’s funny’’ or ``Oh, cool.''
    \end{itemize}
  \item \textbf{D (Differing Perspective):} Offers a different or opposing view.
  \item \textbf{E (Elicit Response):} Seeks info, feedback, or action (e.g., ``What do you think?’’, ``Can you\ldots?’’)
\end{itemize}

\vspace{0.5em}
\textbf{Labeling Instructions:}
\begin{itemize}
  \item For each utterance, output the \textbf{UTTERANCE NUMBER}, \textbf{SPEAKER NAME}, and the ABCDE label(s).
  \item Format examples:
    \begin{itemize}
      \item \texttt{3 Mira [E]} \quad if only E applies to line \#3.
      \item \texttt{2 Oscar []} \quad if no labels apply.
      \item \texttt{1 Jalen [A, C]} \quad for multiple applicable codes.
    \end{itemize}
  \item Use brackets for labels. Separate multiple labels with commas: \texttt{[B, E]}.
  \item You must produce \textbf{exactly \texttt{\{num\_utterances\}}} label lines — no more, no less.
  \item \textbf{No additional text or explanation} should be included.
  \item The line number and speaker must match the transcript exactly.
\end{itemize}

\vspace{0.5em}
Please review the codebook and labeling instructions carefully. Once ready, the full transcript is provided below for labeling.

\vspace{1em}
\texttt{<<<TRANSCRIPT\_START>>>} \\
\texttt{\{Full Transcript\}} \\
\texttt{<<<TRANSCRIPT\_END>>>}

\vspace{0.5em}
There are \texttt{\{num\_utterances\}} utterances in total. Please provide \textbf{EXACTLY \texttt{\{num\_utterances\}} label lines} below.

\subsection{ABCDE - Full Context: Thread Prompt} 

You are a trained qualitative researcher annotating small-group transcripts using the ABCDE Codebook. Your task is to \textbf{label only the final utterance} in the provided transcript segment using the relevant ABCDE codes below. Multiple labels may apply, or none.

\vspace{0.5em}
\textbf{ABCDE Codebook:}
\begin{itemize}
  \item \textbf{A (Agree):} Expresses agreement with another’s idea. Yes/no question responses don't count. Acknowledgments without clear reasoning don’t count either.
  \item \textbf{B (Build on):} Adds information to ideas proposed by someone else. A speaker building on themselves doesn’t count.
  \item \textbf{C (Chat/Comment):}
    \begin{itemize}
      \item \textbf{Chat:} Off-task talk (e.g., life, jokes)
      \item \textbf{Comment:} Reactions like ``That’s funny’’ or ``Oh, cool.''
    \end{itemize}
  \item \textbf{D (Differing Perspective):} Offers a different or opposing view.
  \item \textbf{E (Elicit Response):} Seeks info, feedback, or action (e.g., ``What do you think?’’, ``Can you\ldots?’’)
\end{itemize}

\vspace{0.5em}
\textbf{Labeling Instructions:}
\begin{itemize}
  \item For each utterance, output the \textbf{UTTERANCE NUMBER}, \textbf{SPEAKER NAME}, and the applicable ABCDE labels.
  \item Format examples:
  \begin{itemize}
    \item \texttt{1 Jalen [A]}
    \item \texttt{2 Oscar []}
    \item \texttt{3 Mira [C, E]}
  \end{itemize}
  \item If no labels apply, use empty brackets: \texttt{[]}.
  \item Separate multiple labels with commas inside the brackets: \texttt{[B, E]}.
  \item You must produce \textbf{exactly \texttt{\{num\_utterances\}} label lines} — no more, no less.
  \item \textbf{Do not include any extra text or explanation.}
  \item Line numbers and speaker names must match the transcript \textbf{exactly}.
\end{itemize}

\vspace{0.5em}
Please review the codebook and instructions carefully. Then I will provide the transcript for ABCDE labeling.

\vspace{1em}
Here is a \textbf{threaded transcript} with timestamp, speaker tags, and utterance indices. This is from a group of high school students working on a data science task. An AI agent named Oscar is also present. Oscar’s responses may lag due to voice technology latency — so he may be responding to utterances from several turns ago. Be mindful of this when labeling ABCDE codes.

\vspace{1em}
\texttt{<<<TRANSCRIPT\_START>>>} \\
\texttt{\{Full Treaded Transcript\}} \\
\texttt{<<<TRANSCRIPT\_END>>>}

\vspace{0.5em}
There are \texttt{\{num\_utterances\}} utterances in total. Please provide \textbf{EXACTLY \texttt{\{num\_utterances\}} label lines} below.

\subsection{ABCDE Prompt Based On \cite{martinenghi2024llms}:} 

You will be given a dialogue from a small group of high school students collaborating on a data science project with an AI agent named Oscar. Your task is to \textbf{predict the class of the utterance}.

\vspace{0.5em}
\textbf{The dialogue:}
\begin{flushleft}
\texttt{<<<TRANSCRIPT\_START>>>} \\
\texttt{\{Full Transcript\}} \\
\texttt{<<<TRANSCRIPT\_END>>>}
\end{flushleft}

\vspace{0.5em}
It is essential to consider:
\begin{itemize}
  \item The content of what each speaker says.
  \item The intent behind their speech.
  \item The order of the utterances.
\end{itemize}

\vspace{0.5em}
\textbf{Admissible Classes: Definitions and Examples}
\begin{itemize}
  \item \textbf{A (Agree):} Expresses agreement with another speaker.
    \begin{itemize}
      \item Greta: I think that we're done? \\
            Katie: Okay, good. \quad \texttt{(Katie = A)}
      \item Ana: \ldots you know? \\
            Ben: Yeah. \quad \texttt{(Ben = A)}
    \end{itemize}

  \item \textbf{B (Build on):} Builds on or extends another’s idea.
    \begin{itemize}
      \item Greta: Video games actually bring up the GPA average. \\
            James: Contrary to popular belief. \quad \texttt{(James = B)}
      \item Katie: I think we're all just tall. \\
            Greta: I'm 5'4\ldots \\
            Katie: I'm five nine\ldots \quad \texttt{(Greta and Katie = B)}
    \end{itemize}

  \item \textbf{C (Social Chat/Comments):} Social talk or comments.
    \begin{itemize}
      \item Greta: You're allowed to take it in junior year\ldots \\
            James: What grade are you in? \\
            Greta: We're both juniors. \quad \texttt{(All = C)}
      \item Oscar: Same old sheets. \\
            Greta: It's kind of funny. \quad \texttt{(Greta = C)}
    \end{itemize}

  \item \textbf{D (Differing perspectives):} Offers a different opinion.
    \begin{itemize}
      \item Greta: I think we're done? \\
            Katie: Okay, good. \\
            Oscar: Maybe. But did anyone double-check\ldots \quad \texttt{(Oscar = D)}
      \item Oscar: Napoli's like the pizza place. \\
            Greta: I don't think you can eat, Oscar. \quad \texttt{(Greta = D)}
    \end{itemize}

  \item \textbf{E (Elicit Response):} Requests feedback, ideas, or action.
    \begin{itemize}
      \item Greta: I'm 5'4. I don't know about you guys. \quad \texttt{(Greta = E)}
      \item James: What grade are you in? \quad \texttt{(James = E)}
    \end{itemize}
\end{itemize}

\vspace{0.5em}
\textbf{Labeling Instructions:}
\begin{itemize}
  \item For each utterance, output the \textbf{UTTERANCE NUMBER}, \textbf{SPEAKER NAME}, and the predicted ABCDE label(s).
  \item Format:
    \begin{itemize}
      \item \texttt{1 Jalen [A]}
      \item \texttt{2 Oscar []}
      \item \texttt{3 Mira [C, E]}
    \end{itemize}
  \item If no labels apply, write \texttt{[]}.
  \item Separate multiple labels with commas inside brackets.
  \item You must produce \textbf{exactly \texttt{\{num\_utterances\}} label lines}.
  \item \textbf{Do not include any extra text or explanation}.
  \item Line number and speaker must match the transcript exactly.
\end{itemize}

\subsection{ABCDE Prompt Based On \cite{lee2025capturing}:}

\textbf{[Dialogue Context]} \\
A small group of high school students are collaborating on a data science project with an AI agent named Oscar.

You are a teacher who is assessing the students’ collaborative competency. For each utterance, generate who the addressee is. Think through the speaker’s intention before choosing labels.

\vspace{1em}
\texttt{<<<TRANSCRIPT\_START>>>} \\
\texttt{\{Sliding Window Transcript\}} \\
\texttt{<<<TRANSCRIPT\_END>>>}

\vspace{0.5em}
\textbf{[Categories]} \\
Use the ABCDE Codebook to label the target utterance:
\begin{itemize}
  \item \textbf{A (Agree)} – Expresses agreement with another speaker.
  \item \textbf{B (Build on)} – Builds on or extends another’s idea.
  \item \textbf{C (Social Chat/Comments)} – Social talk or comments.
  \item \textbf{D (Differing perspectives)} – Offers a different opinion to someone else.
  \item \textbf{E (Elicit Responses)} – Requests feedback, ideas, or action from others.
\end{itemize}

\vspace{0.5em}
\textbf{[Output Format]} \\
For each utterance, use the following structure:
\begin{quote}
\texttt{<utterance\_number> <speaker> [<labels>]}
\end{quote}
Use \texttt{[]} if no code applies. \\
\textbf{Example:}
\begin{quote}
\texttt{1 Mira [B]} \\
\texttt{2 Oscar []} \\
\texttt{3 Greta [A, E]}
\end{quote}

\vspace{0.5em}
\textbf{[Instruction]} \\
What category does the following utterance fall into?

\textbf{Target Utterance:}
\begin{flushleft}
\texttt{<<<TARGET\_START>>>} \\
\texttt{f"\{Speaker Name\} \{Transcript\}"} \\
\texttt{<<<TARGET\_END>>>}
\end{flushleft}

\vspace{0.5em}
\textbf{[Example Cases]}

\begin{itemize}
  \item \textbf{A (Agree)}
    \begin{itemize}
      \item Greta: I think that we're done? \\
            Katie: Okay, good. \quad \texttt{(Katie = A)}
      \item Ana: \ldots you know? \\
            Ben: Yeah. \quad \texttt{(Ben = A)}
    \end{itemize}
    \textbf{Non-examples:}
    \begin{itemize}
      \item ``Yeah'' to a yes/no question like ``Did you guys finish?'' → not A.
      \item Simultaneous ``yeahs'' without clear referent → not A.
    \end{itemize}

  \item \textbf{B (Build on)}
    \begin{itemize}
      \item Greta: Video games actually bring up the GPA average. \\
            James: Contrary to popular belief. \quad \texttt{(James = B)}
      \item Katie: I think we're all just tall. \\
            Greta: I'm 5'4\ldots \\
            Katie: I'm five nine\ldots \quad \texttt{(Greta and Katie = B)}
      \item A: Should we go to Cambridge? \\
            B: Yeah, let's go to Harvard Sq. \quad \texttt{(B = B)}
    \end{itemize}
    \textbf{Non-example:}
    \begin{itemize}
      \item A: Where should we go? \\
            B: Harvard Sq. → \textbf{not B} (just a response, not building).
    \end{itemize}

  \item \textbf{C (Social Chat/Comments)}
    \begin{itemize}
      \item Greta: You're allowed to take it in junior year\ldots \\
            James: What grade are you in? \\
            Greta: We're both juniors. \quad \texttt{(All = C)}
      \item Oscar: Same old sheets. \\
            Greta: It's kind of funny. \quad \texttt{(Greta = C)}
    \end{itemize}

  \item \textbf{D (Differing Perspectives)}
    \begin{itemize}
      \item Greta: I think we're done? \\
            Katie: Okay, good. \\
            Oscar: Maybe. But did anyone double-check\ldots \quad \texttt{(Oscar = D)}
      \item Katie: The lower extreme is 60, right? \\
            Greta: Yeah. Or — no, that's not\ldots \quad \texttt{(Greta = A + D)}
      \item Oscar: Napoli's like the pizza place. \\
            Greta: I don't think you can eat, Oscar. \quad \texttt{(Greta = D)}
    \end{itemize}

  \item \textbf{E (Elicit Responses)}
    \begin{itemize}
      \item Greta: I'm 5'4. I don't know about you guys. \quad \texttt{(Greta = E)}
      \item James: What grade are you in? \quad \texttt{(James = B + E)}
      \item Greta: You need to request access again. \quad \texttt{(Greta = E)}
    \end{itemize}
\end{itemize}

\subsection{ABCDE Prompt Based On \cite{qamarllms}:}

You are an intelligent annotator capable of classifying the intention behind each speaker’s utterance. \\
You will be provided with a list of possible Dialogue Acts and their definitions.

You will be given an utterance and your task is to \textbf{predict the correct label for the last utterance only}. \\
You will also be given a snapshot of the conversation to provide context for your prediction.

\vspace{0.5em}
Return the answer in the following format: \\
\texttt{\{line\_number\} \{speaker\} [A, B, ...]} or \texttt{[]} if no codes apply.

\vspace{0.5em}
\textbf{Dialogue Acts and Definitions:}
\begin{itemize}
  \item \textbf{A (Agree):} Expresses agreement with another's idea. Not just replying to a question.
  \item \textbf{B (Build on):} Adds to another's idea with elaboration. Must reference their point.
  \item \textbf{C (Chat/Comment):}
    \begin{itemize}
      \item \textbf{Chat:} Off-task talk (e.g., life, jokes)
      \item \textbf{Comment:} Reactions like ``That's funny'' or ``Oh, cool.''
    \end{itemize}
  \item \textbf{D (Differing Perspective):} Offers a different or opposing view.
  \item \textbf{E (Elicit Response):} Seeks info, feedback, or action (e.g., ``What do you think?'', ``Can you\ldots?'')
\end{itemize}

\vspace{0.5em}
\textbf{Examples:}

\textbf{Example 1:}
\begin{quote}
1 Serena: Okay, I made a new sheet and it's called Scatter Plot. So we're just going to copy and paste the data from the Instagram one and the cumulative GPA one. \\
2 Oscar: Yeah. Box plots are cool for spotting outliers in GPAs like who's the genius and who's chillin? \\
3 Ivy: We're not talking about box plots anymore, Oscar. \\
4 Serena: Something happened. I'll fix it. There we go. \\
5 Ivy: Um, do you guys want to try to take the... I just, I feel bad because I take AP stats, and this is just like, I don't want to just, like, because I know how to do everything at this point. I didn't realize... this is, like, I didn’t realize how much of this is just AP stats. \\
6 Serena: Mhm. \\
7 Ivy: And I did a project on this, like three months ago. \\
8 Serena: Don’t you? Just like, like put... you just, like, copy, copy it, and then you go insert and then you put scatter plot. \\
9 Ivy: Yeah. But then you have to find, um, you have to make sure you have all the labels. Do, uh, do the whole thing, like from here to here. \\
10 Serena: What whole thing? \\
\textbf{Output:} \texttt{10 Serena [E]}
\end{quote}

\textbf{Example 2:}
\begin{quote}
1 Skyler: Oh, sorry. \\
2 Juliette: You can say it's fine. \\
3 Skyler: 64.994, blah blah blah blah blah. \\
4 Eden: Wait what's the next number? \\
5 Skyler: 11765 \\
6 Eden: It was the same one as the older version. \\
7 Juliette: Are you serious right now? \\
8 Eden: Yeah. \\
9 Skyler: What does he mean it's weird. \\
10 Eden: Yeah. What does he mean by that? I don't get it. \\
\textbf{Output:} \texttt{10 Eden [E]}
\end{quote}

\texttt{<<<TRANSCRIPT\_START>>>} \\
\texttt{\{Sliding Window Transcript\}} \\
\texttt{<<<TRANSCRIPT\_END>>>}

\section*{Declaration of Generative AI Software Tools in the Writing Process}

\emph{During the preparation of this work, the author(s) used OpenAI's GPT models  in the following sections: (1) for conducting and generating responses in the large language model (LLM) experiments described; and (2) for drafting and editing portions of the manuscript to improve clarity and style. After using this tool, the authors reviewed and edited the content as needed and takes full responsibility for the content of the publication.}


\end{document}